\pdfoutput=1
\documentclass[letterpaper]{article} 
\usepackage{aaai22}  
\usepackage{times}  
\usepackage{helvet}  
\usepackage{courier}  
\usepackage[hyphens]{url}  
\usepackage{graphicx} 
\usepackage{natbib}  
\usepackage{caption} 
\DeclareCaptionStyle{ruled}{labelfont=normalfont,labelsep=colon,strut=off} 
\usepackage{subfigure}
\usepackage{amsmath}
\usepackage{multirow}
\usepackage{array}
\usepackage{mathtools}
\usepackage{bbm}
\usepackage{algorithm}
\usepackage{algpseudocode}
\usepackage{amsfonts}
\usepackage{booktabs}
\usepackage{makecell}
\usepackage{enumitem} 
\usepackage{url}
\usepackage{color,xcolor}
\usepackage{soul}
\usepackage{stfloats}
\usepackage[switch]{lineno} %
\definecolor{green}{RGB}{0,204,0}

\algtext*{EndWhile}
\algtext*{EndIf}
\algtext*{EndFor}

\usepackage{newfloat}
\usepackage{listings}
\floatstyle{ruled}
\newfloat{listing}{tb}{lst}{}
\floatname{listing}{Listing}
\title{Git: Clustering Based on Graph of Intensity Topology}
\usepackage{bibentry}

\title{Git: Clustering Based on Graph of Intensity Topology}
\author {
    Zhangyang Gao,\textsuperscript{\rm 1 2}
    Haitao Lin, \textsuperscript{\rm 1 2}
    Cheng Tan \textsuperscript{\rm 1 2}
    Lirong Wu \textsuperscript{\rm 1 2}
    Stan. Z Li \textsuperscript{\rm 1}
}
\affiliations {
    \textsuperscript{\rm 1} AI Research and Innovation Lab, Westlake University\\
    \textsuperscript{\rm 2} Zhejiang University\\
    gaozhangyang@westlake.edu.cn, linhaitao@westlake.edu.cn, tancheng@westlake.edu.cn, wulirong@westlake.edu.cn, Stan.ZQ.Li@westlake.edu.cn
}

\begin{document}

\maketitle

\begin{abstract}

\textbf{A}ccuracy, \textbf{R}obustness to noises and scales, \textbf{I}nterpretability, \textbf{S}peed, and \textbf{E}asy to use (ARISE) are crucial requirements of a good clustering algorithm. However, achieving these goals simultaneously is challenging, and most advanced approaches only focus on parts of them. Towards an overall consideration of these aspects, we propose a novel clustering algorithm, namely GIT (Clustering Based on \textbf{G}raph of \textbf{I}ntensity \textbf{T}opology). GIT considers both local and global data structures: firstly forming local clusters based on intensity peaks of samples, and then estimating the global topological graph (topo-graph) between these local clusters. We use the Wasserstein Distance between the predicted and prior class proportions to automatically cut noisy edges in the topo-graph and merge connected local clusters as final clusters. Then, we compare GIT with seven competing algorithms on five synthetic datasets and nine real-world datasets. With fast local cluster detection, robust topo-graph construction and accurate edge-cutting, GIT shows attractive ARISE performance and significantly exceeds other non-convex clustering methods. For example, GIT outperforms its counterparts about $10\%$ (F1-score) on MNIST and FashionMNIST. Code is available at \color{red}{https://github.com/gaozhangyang/GIT}.

\end{abstract}

\section{Introduction}
With the continuous development of the past 90 years \cite{driver1932quantitative, zubin1938technique, tryon1939cluster}, numerous clustering algorithms \cite{jain1999data, saxena2017review, gan2020data} have promoted scientific progress in various fields, such as biology, social science and computer science. As to these approaches, \textbf{A}ccuracy, \textbf{R}obustness, \textbf{I}nterpretability, \textbf{S}peed, and \textbf{E}asy to use (ARISE) are crucial requirements for wide usage. However, most previous works only show their superiority in certain aspects while ignoring others, leading to sub-optimal solutions. How to boost the overall ARISE performance, especially the accuracy and robustness is the critical problem this paper try to address.

Existing clustering methods, e.g., center-based, spectral-based and density-based, cannot achieve satisfactory ARISE performance. Typical center-based methods such as \textit{k}-means \cite{steinhaus1956division, lloyd1982least} and \textit{k}-means++ \cite{arthur2006k, lattanzi2019better} are fast, convenient and interpretable, but the resulted clusters must be convex and depend on the initial state. In the non-convex case, elegant spectral clustering \cite{dhillon2004kernel} finds clusters by minimizing the edge-cut between them with solid mathematical basis. However, it is challenging to calculate eigenvectors of the large and dense similarity matrix and handle noisy or multiscale data for spectral clustering \cite{nadler2006fundamental}. Moreover, the sensitivity of eigenvectors to the similarity matrix is not intuitive \cite{meila2016spectral}, limiting its interpretability. A more explainable way to detect non-convex clusters is density-based method, which has recently attracted considerable attention. Density clustering relies on the following assumption: \textit{data points tend to form clusters in high-density areas, while noises tend to appear in low-density areas}. For example, DBSCAN \cite{ester1996density} groups closely connected points into clusters and leaves outliers as noises, but it only provides \textit{flat} labeling of samples; thus, HDBSCAN \cite{campello2013density, campello2015hierarchical, mcinnes2017accelerated} and DPA \cite{d2021automatic} are proposed to identify hierarchical clusters. In practice, we find that DBSCAN, HDBSCAN, and DPA usually treat overmuch valid points as outliers, resulting in the label missing issue\footnote{Most valid points are identified as noise without proper labels.}. In addition, Mean-shift, ToMATo, FSFDP and their derivatives \cite{comaniciu2002mean, chazal2013persistence, rodriguez2014clustering, ezugwu2021automatic} are other classic density clustering algorithms with sub-optimum accuracy. Recently, some delicate algorithms appear to improve the robustness to data scales or noises, such as RECOME \cite{geng2018recome}, Quickshift++ \cite{jiang2018quickshift++} and SpectACI \cite{hess2019spectacl}. However, the accuracy gain of these methods is limited, and the computational cost increases sharply on large-scale datasets.  Another promising direction is to combine deep learning with clustering \cite{hershey2016deep,caron2018deep,wang2018alternative,zhan2020online}, but these methods are hard to use, time-consuming, and introduce the randomness of deep learning. In summary, as to clustering algotirhms, there is still a large room for improving ARISE performance.

To improve the overall ARISE performance, we propose a novel algorithm named GIT (\textit{Clustering Based on \textbf{G}raph of \textbf{I}ntensity \textbf{T}opology}), which contains two stages: finding local clusters, and merging them into final clusters. We detect locally high-density regions through an intensity function and collect internal points as local clusters. Unlike previous works, we take local clusters as basic units instead of sample points and further consider connectivities between them to consititude a topo-graph describing the global data structure. We point out that the key to improving the \textbf{accuracy} is cutting noisy edges in the topo-graph. Differ from threshold-based edge-cutting, we introduce a knowledge-guidied algorithm to filter noisy edges by using prior class proportion, e.g., 1:0.5:0.1 for a dataset with three unbalanced classes. This algorithm enjoys two advantages. Firstly, it is relatively robust and easy-to-tune when only the number of classes is known, in which case we set the same sample number for all classes. Secondly, it is promising to solve the problem of unbalanced sample distribution given the actual proportion. Treat the balanced proportion as prior knowledge, there is only one parameter ($k$ for kNN searching) need to be tuned in GIT, which is relative \textbf{easy to use}. We further study the \textbf{robustness} of various methods to data shapes, noises and scales, where GIT significantly outperform competitors. We also \textbf{speed} up GIT and its time complexity is $\mathcal{O}(d_s n\log(n))$, where $d_s$ is the dimension of feature channels and $n$ is the number of samples. Finally, we provide visual explanations of each step for GIT to show its \textbf{interpretability}.

In summary, we propose GIT to achieve better ARISE performance, considering both local and global data structures. Extensive experiments confirm this claim.

\begin{figure*}[ht]
	\centering
	    \includegraphics[width=7in]{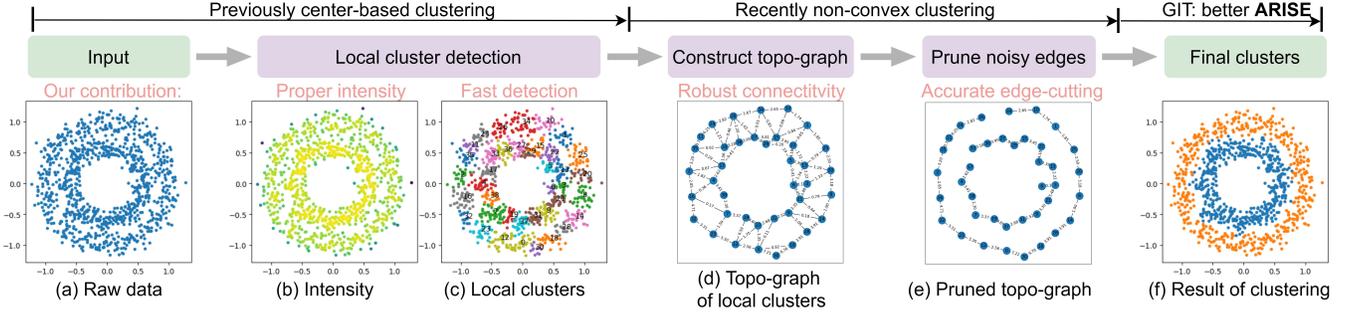}
	    \caption{The pipeline of GIT: after the intensity function in (b) has been estimated from the raw data (two rings) in (a), local clusters (c) are detected through the intensity growing process, seeing Section.~\ref{sec:local_cluster}. Due to the limited number of colors, different local clusters may share the same color. (d) shows that each local cluster is represented as a vertex in the topological graph, and edges between them are calculated. In (e), noisy edges are pruned and the real global structure (two rings) is revealed. Finally, the clustering task is finished by merging connected local lusters, as shown in (f).}
	    \label{fig:pipeline}
\end{figure*}

\begin{table}[H]
  \centering
  \caption{Commonly used symbols}
  \resizebox{0.9 \columnwidth}{!}{
  \begin{tabular}{cc}
  \toprule
       Symbol & Description \\ \midrule
       $\mathcal{X}$ & Dataset, containing $\boldsymbol{x}_1,\boldsymbol{x}_2,\ldots,\boldsymbol{x}_n$. \\

       $n$ & The number of samples.\\
       $k$ & The hyper-parameter, used for kNN searching.\\
       $d_s$      & The number of input feature dimension.\\
       $r_i$                &  The root (or intensity peak) of the $i$-th local cluster.\\
       $R(\boldsymbol{x})$  &  The root of the local cluster containing $\boldsymbol{x}$.\\
       $\mathcal{R}$      & The set of root points.\\
       $S(\cdot,\cdot)$ & Similarity between local clusters.\\
       $s_l(\cdot,\cdot)$ & Similarity between samples based on local clusters.\\
       $f(\boldsymbol{x})$ &  The intensity of $\boldsymbol{x}$ \\
       $L_{\lambda}^+$ & The super-level set of the intensity function with threshold $\lambda$. \\
       $C_{i}(t)$ & The $i$-th local cluster at time $t$.\\
  \bottomrule
  \end{tabular}
  }
  \label{tab:symbols}
\end{table}

\section{Method}
\subsection{Motivation, Symbols and Pipeline }
\begin{figure}[h]
    \centering
    \includegraphics[width=2.5in]{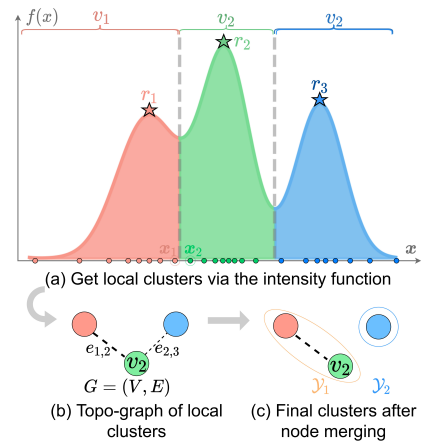}
    \caption{Motivation with 1D data. In (a), we estimate the intensity function $f(x)$ from samples and partition data into local clusters $\{v_1, v_2, v_3\}$ along valleys. Points with the largest intensity are called \textit{roots} of local clusters, such as $r_1$, $r_2$ and $r_3$. ($\boldsymbol{x}_1$, $\boldsymbol{x}_2$) is a boundary pair connecting $v_1$ and $v_2$. (b) shows the topology graph containing local clusters (nodes) and their connectivity (edges), where $e_{1,2}$ is stronger than $e_{2,3}$. In (c), by cutting $e_{2,3}$, we get final clusters $\mathcal{Y}_1=\{v_1, v_2\}$ and $\mathcal{Y}_2=\{v_3\}$.}
    \label{fig:motivation_1d}
\end{figure}

In Fig.~\ref{fig:motivation_1d}, we illustrate the motivation that local clusters are point sets in high-intensity regions and they are connected by shared boundaries. And GIT aims to

\begin{enumerate}[itemindent=1em]
    \item determine local clusters via the intensity function;
    \item construct the connectivity graph of local clusters;
    \item cut noisy edges for reading out final clusters, each of which contains several connected local clusters.
\end{enumerate}

The commonly used symbols and the overall pipeline are shown in Table.~\ref{tab:symbols} and Fig.~\ref{fig:pipeline}.

\subsection{Intensity Function}
To identify local clusters, non-parametric kernel density estimation (KDE) is employed to estimate the data distribution. However, the kernel-based \cite{parzen1962estimation, davis2011remarks} or k-Nearest Neighborhood(KNN)-based \cite{loftsgaarden1965nonparametric} KDE either suffers from global over-smoothing or local oscillatory \cite{yip2006dynamic}, both of which are not suitable for GIT. Instead, we use a new intensity function $f(\boldsymbol{x})$ for better empirical performance:
\begin{align}
\label{eq:intensity}
\normalsize{
    f(\boldsymbol{x}) = \frac{1}{|\mathcal{N}_{\boldsymbol{x}}|}\sum_{\boldsymbol{y} \in \mathcal{N}_{\boldsymbol{x}}}{e^{-d(\boldsymbol{x},\boldsymbol{y})}};d(\boldsymbol{x},\boldsymbol{y}) = \sqrt{\sum_{j=1}^{s}\frac{(x_j-y_j)^2}{\sigma_j^2}},
}%
\end{align}
where $\mathcal{N}_{\boldsymbol{x}}$ is $\boldsymbol{x}$'s neighbors for intensity estimation and $\sigma_j $ is the standard deviation of the $j$-th dimension. With the consideration of $\sigma_j$, $d(\cdot,\cdot)$ is robust to the absolute data scale. We estimate the intensity for each point in its neighborhood, avoiding the global over-smoothing. The exponent kernel is helpful to avoid local oscillatory, and we show that $f(\boldsymbol{x})$ is Lipschitz continuous under a mild condition, seeing Appendix.~\ref{sup:idensity_function}.

\subsection{Local Clusters}
\label{sec:local_cluster}
A local cluster is a set of points belonging to the same high-intensity region. Once the intensity function is given, how to efficiently detect these local clusters is the key problem.

\begin{figure}[h]
    \centering
    \includegraphics[width=2in]{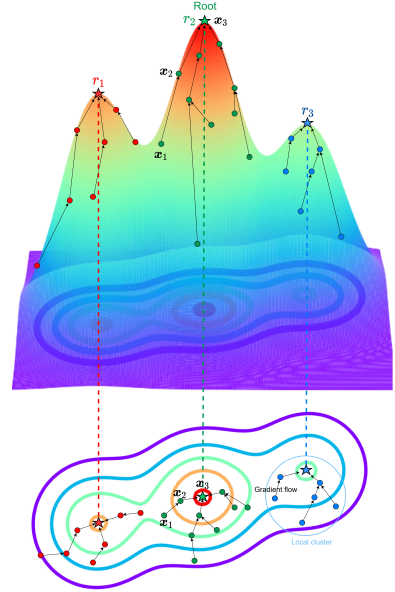}
    \caption{ Algorithm realization of intensity growth process. Several data points and the intensity contour are presented. As the intensity threshold gradually decreases from 1 to 0, points with higher intensity appear earlier. For a newborn point $\boldsymbol{x}_i$, if there is a neighbor $\boldsymbol{x}_j$ with the greatest gradient starting at $\boldsymbol{x}_i$ and $f(\boldsymbol{x}_j)\geq f(\boldsymbol{x}_i)$, we connect $\boldsymbol{x}_j$ with $\boldsymbol{x}_i$, and call $\boldsymbol{x}_j$ as the parent of $\boldsymbol{x}_i$.Otherwise, there is no valid parent for $\boldsymbol{x}_i$, and we treat $\boldsymbol{x}_i$ as the root of a new local cluster. Finally, points sharing the same root form local clusters.}
    \label{fig:3d_find_peak}
\end{figure}

We introduce a fast algorithm that collects points within the same intensity peak by searching along the gradient direction of the intensity function (refer to Fig.~\ref{fig:3d_find_peak}, Appendix.~\ref{sup:idensity_function}). Although similar approaches have been proposed before \cite{chazal2013persistence, rodriguez2014clustering}, we provide comprehensive time complexity analysis and take \textit{connected boundary pairs} (introduced in Section.~\ref{sec:construct_topo_graph}) of adjacent local clusters into considerations. To clarify this approach, we formally define the \textit{root}, \textit{gradient flow} and \textit{local clusters}. Then we introduce a \textit{density growing process} for an efficient algorithmic implementation.

\paragraph{Definition.} (\textit{root}, \textit{gradient flow}, \textit{local clusters})
 Given intensity function $f(\boldsymbol{x})$, the set of \textit{roots} is $\mathcal{R}=\{ \boldsymbol{r}| \nabla f(\boldsymbol{r})=0, |\nabla^2 f(\boldsymbol{r})|<0 , \boldsymbol{r} \in \mathcal{X} \}$, i.e., the local maximum. 
 For any point $\boldsymbol{x} \in \mathbb{R}^{d_s}$, there is a \textit{gradient flow} $\pi_{\boldsymbol{x}}:[0,1] \mapsto \mathbb{R}^{d_s}$, starting at $\pi_{\boldsymbol{x}}(0)=\boldsymbol{x}$ and ending in  $\pi_{\boldsymbol{x}}(1)=R(\boldsymbol{x})$, where $R(\boldsymbol{x}) \in \mathcal{R}$. A \textit{local cluster} a set of points converging to the same root along the \textit{gradient flow}, e.g., $\{\boldsymbol{x}|R(\boldsymbol{x}) \in \mathcal{R},\boldsymbol{x} \in \mathcal{X}\}$. To better understand, please see Fig.~\ref{fig:3d_find_peak} in Appendix.~\ref{sup:idensity_function}.\\

\begin{figure}[h]
    \centering
    \includegraphics[width=3.3in]{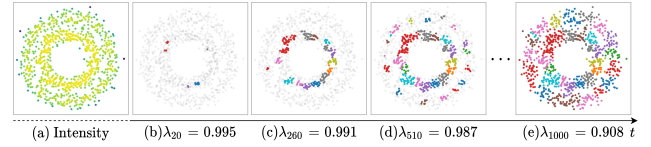}
    \caption{ Example of the intensity growth process. (a) shows the estimated intensities; and (b), (c), (d), (e) show the growing process of local clusters as $\lambda_t$ gradually decays. We mark the unborn points in light gray and highlight local clusters. Limited by the number of colors, different local clusters may share the same color. }
    \label{fig:density_growing_process}
\end{figure}

\paragraph{Intensity Growing Process.} As shown in Fig.~\ref{fig:density_growing_process}, points with higher intensities appear earlier and local clusters grow as the time goes on. We formally define the intensity growing process via a series of super-level sets w.r.t. $\lambda$:
\begin{equation}
    L_{\lambda}^{+}=\{\boldsymbol{x}| f(\boldsymbol{x}) \geq \lambda \}.
\end{equation}
We introduce a time variable $t, 1 \leq t \leq n$, such that $\lambda_{t-1} \geq \lambda_{t}$. At time $t$, the $i$-th local cluster is 
\begin{equation}
    C_{i}(t)=\{ \boldsymbol{x}| R(\boldsymbol{x})=\boldsymbol{r}_i, \boldsymbol{x} \in L_{\lambda_t}^{+} \},
\end{equation}
where $R(\cdot)$ is the root function, $\boldsymbol{r}_i$ is the $i$-th root. It is obvious that $L_{\lambda_{t-1}}^{+} \subset L_{\lambda_{t}}^{+}$ and $L_{\lambda_{n}}^{+}=\mathcal{X}$. When $t=n$, we can get all local clusters $C_{1}(n), C_{2}(n),\ldots$. Next, we introduce a recursive function to efficiently compute $C_{i}(t)$.

\paragraph{Efficient Implementation.} For each point $\boldsymbol{x}_i$, to determine whether it creates a new local cluster or belongs to existing one, we need to compute its intensity\footnote{The intensity can be obtained by searching kNN and applying Eq.~\ref{eq:intensity} with time complexity $\mathcal{O}(d_s n \log{n})$ (using kd-tree).} and its parent point $Pa(\boldsymbol{x}_i)$ along the gradient direction. We determine $Pa(\boldsymbol{x}_i)$ as the neighbor which has maximum directional derivative from $\boldsymbol{x}_i$ to $Pa(\boldsymbol{x}_i)$, that is
\begin{align}
    \label{eq:Prx}
    Pa(\boldsymbol{x}_i)={\arg\max}_{\boldsymbol{x}_p\in L_{f(\boldsymbol{x}_i)}^{+} \cap \mathcal{N}_{\boldsymbol{x}_i}}\frac{f(\boldsymbol{x}_p)-f(\boldsymbol{x}_i)}{||\boldsymbol{x}_p-\boldsymbol{x}_i||},
\end{align}
where $\mathcal{N}_{\boldsymbol{x}_i}$ is the neighborhood system of $\boldsymbol{x}_i$. We sort and re-index all samples by their intensities. With a slight abuse of notation, the sorted $\{\boldsymbol{x}_1,\boldsymbol{x}_2,\ldots,\boldsymbol{x}_n\}$ satisfies $f(\boldsymbol{x}_{i}) \geq f(\boldsymbol{x}_{i+1})$ for all $i$. If $\lambda_{t-1}, L_{\lambda_{t-1}}^{+}$ and $C_{i}(t-1)$ are known, we can get $\lambda_{t}, L_{\lambda_{t}}^{+}$ and $C_{i}(t)$ following
\begin{equation}
    \begin{cases}
        \lambda_{t}^{+} = f(\boldsymbol{x}_t)\\
        L_{\lambda_{t}}^{+} = L_{\lambda_{t-1}}^{+} + \{\boldsymbol{x}_t\}, \\
    \end{cases}
    \label{eq: local_cluster_recursive1}
\end{equation}
and
\begin{equation}
    \small{
    C_{i}(t)=
    \begin{cases}
    \{ \boldsymbol{x}_t \} & i = n_{t-1}, Pa(\boldsymbol{x}_t) = \boldsymbol{x}_t\\
    
    C_{i}(t-1) + \{ \boldsymbol{x}_t \} & R(Pa(\boldsymbol{x}_t)) = \boldsymbol{r}_i\\
    
    C_{i}(t-1) & \mathrm{else,}\\
    \end{cases}
    }%
    \label{eq: local_cluster_recursive2}
\end{equation}
where $i=1,2,\ldots, n_{t-1}$\footnote{$n_{t-1}$ is the minimum index of empty local clusters at $t-1$: $|C_i(t-1)|>0$, if $i<n_{t-1}$; $|C_i(t-1)|=0$, if $i \geq n_{t-1}$}. Eq.~\ref{eq: local_cluster_recursive2} means that 1) if $\boldsymbol{x}_i$'s father is itself, $\boldsymbol{x}_i$ is the peak point and creates a new local cluster $\{\boldsymbol{x}_i\}$, 2) if $\boldsymbol{x}_i$'s father shares the same root with $C_{i}(t-1)$, $\boldsymbol{x}_i$ will be absorbed by $C_{i}(t-1)$ to generate $C_{i}(t)$, and 3) local clusters which is irrelevant to $\boldsymbol{x}_i$ remain unchanged. We treat $R(\cdot)$ as a mapping table and update it on-the-fly. $C_{i}(t)$ is obviously recursive \cite{davis2013computability}. The time complexity of computing $C_i(n)$ is $\mathcal{O}(n\times  \text{cost}(Pa))$\footnote{$\text{cost}(Pa)$ is complexity of computing function $Pa(\cdot)$.}, where $\text{cost}(Pa)$ is $\mathcal{O}( k d_s \log n)$ for querying $k$ neighbors.

Note that the initial state is 
\begin{equation}
    \begin{cases}
      \lambda_{0} = 1 \\
      L_{\lambda_{0}}^{+}=\phi\\
      C_{i}(0)=\phi, i=1.\\
    \end{cases}
    \label{eq: local_cluster_init}
\end{equation}

In summary, by applying Eq.~\ref{eq:Prx}-\ref{eq: local_cluster_init}, we can obtain local clusters $C_{1}(n), C_{2}(n),\ldots$ with time complexity $\mathcal{O}(d_s n \log{n})$. For simplicity, we write $C_{i}(n)$ as $C_{i}$ in later sections. The detailed local cluster detection algorithm \ref{algorithm: local_cluster_detecting} can be found in the Appendix.

\subsection{Construct Topo-graph}
\label{sec:construct_topo_graph}
Instead of using local clusters as final results \cite{comaniciu2002mean, chazal2013persistence, rodriguez2014clustering, jiang2018quickshift++}, we further consider connectivities between them for improving the results. For example, local clusters that belong to the same class usually have stronger connectivities. Meanwhile, there is a risk that noisy connectivities may damage performance. We manage local clusters and their connectivities with a graph, namely topo-graph, and the essential problem is cutting noisy edges.

\paragraph{Connectivity.} According to UPGMA standing for \textit{unweighted pair group method using arithmetic averages} \cite{10.5555/46712, gan2020data}, the similarity $S(\cdot,\cdot)$ between $C_{i}, C_{j} \cup C_{k}$ can be obtained from the Lance-Williams formula:
\begin{equation}
    \begin{split}
         S(C_{i}, C_{j} \cup C_{k}) = \quad \quad \quad \quad  \quad \quad \quad \quad \quad  \quad \quad \quad \quad\\
        \frac{|C_{j}|}{|C_{j}|+|C_{k}|} S(C_{i}, C_{j}) + \frac{|C_{k}|}{|C_{j}|+|C_{k}|}S(C_{i}, C_{k}),
    \end{split}
    \label{eq: distance_local_clusters1}
\end{equation}
and 
\begin{equation}
     S(C_{i}, C_{j}) = \frac{1}{|C_{i}||C_{j}|} \sum_{\boldsymbol{x} \in C_{i},\boldsymbol{y} \in C_{j}}{s_l(\boldsymbol{x},\boldsymbol{y})},
    \label{eq: distance_local_clusters2}
\end{equation}
where Eq.~\ref{eq: distance_local_clusters2} can be derived from Eq.~\ref{eq: distance_local_clusters1}. We define the similarity between points as:
\begin{equation}
    s_l(\boldsymbol{x},\boldsymbol{y})=
    \begin{cases}
      e^{-||\boldsymbol{x}-\boldsymbol{y}||} & \boldsymbol{x} \in \mathcal{N}_{\boldsymbol{y}},\boldsymbol{y} \in \mathcal{N}_{\boldsymbol{x}}, R(\boldsymbol{x}) \neq R(\boldsymbol{y})\\
      0 & \mathrm{else}.
    \end{cases}
    \label{eq: point_distance}
\end{equation}
Note that $s_l(\boldsymbol{x},\boldsymbol{y})$ is non-zero only if $\boldsymbol{x}$ and $\boldsymbol{y}$ are mutual neighborhood and belong to different local clusters, in which case $(\boldsymbol{x},\boldsymbol{y})$ is a \textit{boundary pair} of adjacent local clusters, e.g., $\boldsymbol{x}_1$ and $\boldsymbol{x}_2$ in Fig.~\ref{fig:motivation_1d}. Fortunately, all the boundary pairs among local clusters can be obtained from previous intensity growing process, without further computation.

\paragraph{Topo-graph.} We construct the topo-graph as $G=(V, E)$, where $v_i$ is the $i$-th local cluster $C_{i}$ and $e_{i, j}=S(C_{i}, C_{j})$.

\quad

In summary, we introduce a well-defined connectivity between local clusters using connected boundary pairs. The detailed topo-graph construction algorithm \ref{algorithm: topo-graph construction and pruning} can be found in the Appendix, whose time complexity is $\mathcal{O}(kn)$.

\subsection{Edges Cutting and Final Clusters}
To capture global data structures, connected local clusters are merged as final clusters, as shown in Fig.~\ref{fig:pipeline}(e, f). However, the original topo-graph is usually dense, indicating redundant (or noisy) edges which lead to trivial solutions, e.g., all the local clusters are connected, and there is one final cluster. In this case, cutting noisy edges is necessary and the simplest way is to set a threshold to filter weak edges or using spectral clustering. However, we find that these methods are either sensitive, hard to tune or inaccurate, which severely limits the widely usage of similar approaches. Is there a robust and accurate way to filter noisy edges using available prior knowledge?

\paragraph{Auto Edge Filtering.}  To make GIT more easy to use and accurate, we automatically filter edges with the help of pior class proportion. Firstly, we define a \textit{metric about final clusters} by comparing the predicted and pior proportions, such that the higher the proportion score, the better the result (seeing the next paragraph). We sort edges with connectivity strengths from high to low, in which order we merge two end local clusters for each edge if this operation gets a higher score. In this process, edges that cannot increase the metric score are regarded as noisy edges and will be cutted.

\paragraph{Metric about Final Clusters.} Let $\boldsymbol{p}=[p_1, p_2,\ldots, p_m]$ and  $\boldsymbol{q}=[q_1, q_2,\ldots, q_{m'}]$ be predicted and predefined class proportions, where $\sum_{i=1}^{m}{p_i}=1, \sum_{i}^{m'}{q_i}=1$. We take the similarity between $\boldsymbol{p}$ and $\boldsymbol{q}$ as the aforementioned metric score. Because $m$ and $m'$ may not be equal and the order of elements in $\boldsymbol{p}$ and $\boldsymbol{q}$ is random, it is not feasible to use conventional Lp-norm. Instead, we use Wasserstein Distance to measure the similarity and define the metric $\mathcal{M}(\boldsymbol{p},\boldsymbol{q})$ as:
\begin{equation}
\label{eq:Wasserstein}
\begin{aligned}
  &  \mathcal{M}(\boldsymbol{p},\boldsymbol{q}) = \text{Exp}(- \underset{\gamma_{i, j} \geq 0}{min}\sum_{i, j} \gamma_{i, j})\\
  & s.t. 
  \quad \quad \sum_{j} \gamma_{i, j} = p_i; \quad  \sum_{i} \gamma_{i, j} = q_j,\\
\end{aligned}
\end{equation}
where the original Wasserstein Distance is  $\underset{\gamma_{i, j} \geq 0}{min}\sum_{i, j} \gamma_{i, j} d_{i, j}$, $\gamma_{i, j}$ is the transpotation cost from $i$ to $j$ , and we set $d_{i, j}=1$. Because $\mathrm{dim}(p_i)=\mathrm{dim}(q_j)=1$, Eq.~\ref{eq:Wasserstein} can be efficiently calculated by linear programming.

\quad

In summary, we introduce a new metric of class proportion and use it to guide the process of edge filtering, then merge connected local clusters as final clusters. Compared to threshold-based or spectral-based edge cutting, this algorithm considers available prior knowledge to reduce the difficulty of parameter tuning or provide higher accuracy. By default, we set the same proportion for each category if only the number of classes is known. Besides, it is promising for dealing with sample imbalances if we know the actual proportion, seeing Section.~\ref{exp:accuracy} for experimetal evidence. 

\section{Experiments}

In this section, we conduct experiments on a series of synthetic, small-scale and large-scale datasets to study the \textit{Accuracy}, \textit{Robustness to shapes, noises and scales}, \textit{Speed} and \textit{Easy to use}, while the \textit{Interpretability} of GIT has been demonstrated in previous sections. As an extension, we further investigate how PCA and AE (autoencoder) work with GIT to further improve the accuracy.

\paragraph{Baselines.} GIT is compared to density-based methods, such as FSFDP \cite{rodriguez2014clustering}, HDBSCAN \cite{mcinnes2017accelerated}, Quickshift++ \cite{jiang2018quickshift++}, SpectACl \cite{hess2019spectacl} and DPA \cite{d2021automatic}. Besides, classical Spectral Clustering and \textit{k}-means++ \cite{scikit-learn}\footnote{Some classical algorithms are ignored because they usually perform worse than the latest ones, including  OPTICS \cite{ankerst1999optics}, DBSCAN \cite{ester1996density} and mean-shift \cite{comaniciu2002mean}. Few recent works are also overlooked due to the lack of open-source python code, such as RECOME \cite{geng2018recome} and better \textit{k}-means++ \cite{lattanzi2019better}. }. Since there is little work comprehensively comparing recent clustering algorithms, our results can provide a reliable baseline for subsequent researches. For simplicity, we use these abbreviations: \textbf{FSF} (FSFDP), \textbf{HDB} (HDBSCAN), \textbf{QSP} (QuickshiftPP), \textbf{SA} (SpectACI), \textbf{DPA} (DPA), \textbf{SC} (Spectral Clustering) and \textbf{KM} (\textit{k}-means++).

\paragraph{Metrics.} We report \textbf{F1-score} \cite{Larsen1999}, Adjusted Rand Index (\textbf{ARI}) \cite{Hubert1985} and Normalized Mutual Information (\textbf{NMI}) \cite{vinh2010information} to measure the clustering results. F1-score evaluates both each class's accuracy and the bias of the model, ranging from $0$ to $1$. ARI is a measure of agreement between partitions, ranging from $-1$ to $1$, and if it is less than $0$, the model does not work in the task. NMI is a normalization of the Mutual Information (MI) score, where the result is scaled to range of 0 (no mutual information) and 1 (perfect correlation). Besides, because HDBSCAN and DPA perfer to drop some valid points as noises, we also consider the fraction of samples assigned to clusters, namely \textit{cover rate}. If the cover rate is less than 0.8, we will ignore this result and mark it in gray. The mathmatical formulas of these metrics can be found in the Appendix.~\ref{sup:metrics}. For each methods, we carefully tune the hyperparameters (which can be found in the open-source code), and report the best results.

\paragraph{Datasets.} We evaluate algorithms on synthetic, small-scale and large-scale datasets. In Table.~\ref{Table:datset}, we count the number of samples, feature dimensions, number of classes and balance rate for real-world datasets. The balance rate is the sample ratio of the smallest calss to the largest class. All these datasets are available online \cite{asuncion2007uci,deng2012mnist,xiao2017fashion}. 

\begin{table}[h]
\centering
\caption{Statistics of real-world datasets.}
\resizebox{0.9 \columnwidth}{!}{
\begin{tabular}{cccccc}
    \toprule
        & dataset    & \#samples & \#dim & \#class & balance rate\\
    \midrule
    \multirow{4}{*}{\rotatebox{90}{small}}
    & Iris        & 150   & 4     & 3     & 1.00\\
    & Wine        & 178   & 13    & 3     & 0.68\\
    & Hepatitis   & 154   & 19    & 2     & 0.26\\
    & Cancer      & 569   & 30    & 2     & 0.59\\
    \hline
    \multirow{5}{*}{\rotatebox{90}{large}}
    & Olivetti face   & 400            & 4096     & 40 & 1.00\\
    & MNIST           & 60000        & 784      & 10   & 1.00\\
    & FMNIST           & 60000        & 784      & 10   & 1.00\\
    & Frogs    & 7195          & 22        & 10  & 0.02\\
    & Codon   & 13028      &24     & 11    & 0.01\\
    \bottomrule
    \end{tabular}
}
\label{Table:datset}
\end{table}

\paragraph{Platform.} The platform for our experiment is ubuntu 18.04, with a AMD Ryzen Threadripper 3970X 32-Core cpu and 256GB memory. We fairly compare all algorithms on this  platform for research convenience.

\subsection{Accuracy}
\label{exp:accuracy}
\paragraph{Objective and Setting.} 
To study the accuracy of various algorithms, we: 1) firstly compare GIT with density-based FSF, HDB, QSP, SA, and DPA on small-scale datasets to determine their priority  order, 2) and secondly choose the top-3 (F1-score) density-based algorithms and \textit{k}-means++ as baselines to further compare with GIT on large-scale datasets. With the exception of Frogs and Codon, which are extremely unbalanced, all the class proportions are set to be the same in GIT, e.g. 1:1:1 for three classes.

\paragraph{Result and Analysis.} Comparisons on small-scale and large-scale datasets are shown in Table.~\ref{table:GIT_density_small} and Table.~\ref{table:GIT_density_large}. We notice that GIT outperforms other approaches, with the top-3 ARI, top-2 NMI and top-1 F1-score in all cases. GIT also exceeds competitors up to 6\% and 8\% F1-score in the highly unbalanced case (Frogs and Condon) by specifying the actual prior class proportions. In addition, we find that the performance gains from GIT seem to be negatively correlated with the feature dimension, indicating the potential dimension curse. And we will introduce how to mitigate this problem through dimension reduction in Section.~\ref{sec:dim_reduction_GIT}. Finally, the runtime of GIT is acceptable, especially on large scale datasets, where GIT is much faster than recent density-based clustering methods, such as HDB, QSP and SA.

\begin{table}[h]
\caption{GIT vs density-based algorithms on small-scale datasets. The best and sub-optimum results are highlighted in bold and underline, respectively. Gray results are ignored because their cover rate is less than 0.8.}

\centering
\resizebox{0.9\columnwidth}{!}{
\begin{tabular}{cccccccc}
\toprule
                              &        &DPA & FSF & HDB & QSP & SA & \textbf{GIT}  \\
\midrule
\multirow{3}{*}{\rotatebox{90}{Iris}} 
                                & F1-score & \color{gray}{0.83} &  0.56     & 0.57      & \underline{0.80}      &  0.78    & \textbf{0.88} \\
                              & ARI    & \color{gray}{0.57} &  0.57     & \underline{0.58}      &  0.56       &  0.56    & \textbf{0.71} \\
                              & NMI    & \color{gray}{0.68} &  0.73     & \underline{0.74}      &  0.58        & 0.63     & \textbf{0.76} \\
\midrule
\multirow{3}{*}{\rotatebox{90}{Wine}}        
                                & F1-score & 0.38 & 0.62      & 0.54      &  \underline{0.73}        & 0.70  &  \textbf{0.90}  \\
                              & ARI    & 0.05 &  0.31     & 0.30      &  \underline{0.39}        & 0.36  &  \textbf{0.71}    \\
                              & NMI   & 0.15  & 0.38      & 0.42      &  \underline{0.44}        & 0.36  &  \textbf{0.76}    \\
\midrule
\multirow{3}{*}{\rotatebox{90}{ Hepatitis}} 
                              & F1-score & \color{gray}{0.42} & \underline{0.77}      & 0.71      &  0.71        & 0.72  &  \textbf{0.78}  \\
                              & ARI    & \color{gray}{-0.05} & \textbf{0.50}      & 0.05      &  0.02        & 0.06  &  \underline{0.23}    \\
                              & NMI   & \color{gray}{0.14}  & \textbf{0.46}      & 0.02      &  0.00        & 0.01  &  \underline{0.12}    \\
\midrule
\multirow{3}{*}{\rotatebox{90}{Cancer}}   
                              & F1-score & 0.70 & 0.72      & 0.78      &  0.78        & \underline{0.92}  &  \textbf{0.93}  \\
                              & ARI  & 0.00   & 0.00      & 0.40      &  0.41        & \underline{0.69}  &  \textbf{0.73}    \\
                              & NMI  & 0.00   & 0.00      & 0.34      &  0.34        & \underline{0.57}  &  \textbf{0.65}    \\
\bottomrule
\end{tabular}
}

\label{table:GIT_density_small}
\end{table}

\begin{table}[h]
\caption{GIT vs SOTA algorithms on large-scale datasets.}

\centering
\resizebox{0.9\columnwidth}{!}{
    \begin{tabular}{cccccccc}
    \toprule
                                  &        & KM & SC & HDB & QSP & SA & \textbf{GIT}  \\
    \midrule
    \multirow{4}{*}{\rotatebox{90}{Face}} 
                                    & F1-score & 0.52    & 0.37  &  \color{gray}{0.34}     & \underline{0.60}      & 0.34      &  \textbf{0.62} \\
                                  & ARI    & \underline{0.38}  & 0.19  &  \color{gray}{0.08}     & \underline{0.38}      &  0.21       &  \textbf{0.45}  \\
                                  & NMI    & 0.74  & 0.66  &  \color{gray}{0.61}     & \textbf{0.79}      &  0.61        & \underline{0.78}  \\
                                  & time  & 2.6s & 0.4s    & \color{gray}{0.9s}      & 0.8s      &  1.1s     & 2.1s\\
    \midrule
    \multirow{4}{*}{\rotatebox{90}{MNIST}}        
                                    & F1-score & \underline{0.50}    & 0.41  & \color{gray}{0.99}      & 0.45      &  0.40        & \textbf{0.59}   \\
                                  & ARI    & \underline{0.36}  & 0.33  &  \color{gray}{0.99}     & 0.13      &  0.17        & \textbf{0.42}     \\
                                  & NMI   & \underline{0.45}   & 0.44  & \color{gray}{0.99}      & \underline{0.45}      &  0.33        & \textbf{0.53}  \\
                                  & time  & 76.7s  & 407.7s    & \color{gray}{2037.0s}     & 3384.0s     &   4096s       & 422.1s\\
    \midrule
    \multirow{4}{*}{\rotatebox{90}{FMNIST}} 
                                  & F1-score & 0.39     & 0.43  & \color{gray}{0.06}      & 0.42      &  \underline{0.47}        & \textbf{0.56}  \\
                                  & ARI    & \textbf{0.35}  & \underline{0.34}  & \color{gray}{0.01}      & 0.16      &  0.29        & 0.32   \\
                                  & NMI   & \textbf{0.51}   & \underline{0.49}  & \color{gray}{0.07}      & 0.41      &  0.45        & \textbf{0.51}  \\
                                  & time  & 54.6s    & 397.7s  & \color{gray}{1647.9s}  & 3832.6s  & 4684s   &  444.5s  \\
    \midrule
    \multirow{4}{*}{\rotatebox{90}{Frogs}}   
                                  & F1-score & 0.47     & \underline{0.60}  & \color{gray}{0.95} & 0.50    &  \underline{0.60}        & \textbf{0.66}  \\
                                  & ARI  & 0.40    & 0.41  & \color{gray}{0.96} & 0.21    &  \underline{0.49}        & \textbf{0.69}   \\
                                  & NMI  & \underline{0.61}    & 0.60  & \color{gray}{0.93} & 0.45    &  0.45        & \textbf{0.66}    \\
                                  & time & 0.18   & 3.7s  & \color{gray}{1.2s} & 0.5s & 2.9s & 3.2s\\
    \midrule
    \multirow{4}{*}{\rotatebox{90}{Codon}}   
                                  & F1-score & 0.25     & \underline{0.37}  & \color{gray}{0.21}  & 0.24      &  0.19        & \textbf{0.45}    \\
                                  & ARI  & 0.19    & \underline{0.24}  & \color{gray}{0.05}  & 0.04      &  0.02        & \textbf{0.31}    \\
                                  & NMI  & 0.33    & \underline{0.37}  & \color{gray}{0.24}   & 0.21      &  0.02       & \textbf{0.39}    \\
                                  & time & 1.1s   & 11.6s  & \color{gray}{10.2s} & 7.9s  & 82s & 16s \\
    \bottomrule
    \end{tabular}
}
\label{table:GIT_density_large}
\end{table}

\subsection{Easy to Use}
\label{sec: easy_to_use}
We introduce some experience for parameter tuning and auxiliary tool for data structure analysis.

\paragraph{Parameters Settings.} Apart from the prior class proportion, there is only one hyper-parameter $k$ in GIT for kNN searching, which is usually less than 100. On large-scale datasets, we choose $k$ from [30,40,50,60,70,80,90,100]. When only the number of classes is known, simply setting the ratio of all classes to be the same can generally yield good results. We admit that GIT cannot determine the number of classes from scratch and leave it for future work. If the actual proportion is known, GIT can obtain better results, even in highly unbalanced classes, seeing Frogs and Codon in Table.~\ref{table:GIT_density_large}. 

\paragraph{Topo-graph Visualization.} The topo-graph describes the global structure of the dataset, and GIT uses it to discover final clusters. We also provide API for users to visualize these topo-graphs to understand the inner structure better \cite{d2021automatic}. One of the example of topo-graph can be found in Fig.~\ref{fig:pipeline}(e).

\subsection{Speed}
\paragraph{Objective and Setting.} 
To study the scalability of GIT, we generate artificial data from the mixture of two Gaussian with class proportion 1:1. We examine how dimension ($d_s$), sample number ($n$) and hyper-parameter ($k$) affect the runtime, where $n \in [10^4,10^6]$, $d_s \in [10,10^3]$ and $k \in [10,10^2]$. By default, $d_s=10, n=10^4$ and $k=50$.

\begin{figure}[H]
    \centering
    \includegraphics[width=3.3in]{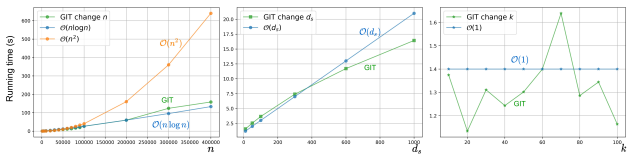}
    \caption{ The actual running time when changing  sample number ($n$), feature dimension ($d_s$)and hyper-parameter ($k$).  }
    \label{fig:speed}
\end{figure}

\paragraph{Result and Analysis.} 
The runtime impact of $n$,$d_s$ and $k$ is shown in Fig.~\ref{fig:speed}. According to previous analysis, the time complexity of GIT is $\mathcal{O}(d_s n \log n)$, which is indeed confirmed by the experimental results. Since $k$ is limited within 100, seeing Section.~\ref{sec: easy_to_use}, its effect on time complexity can be treated as a constant.

\subsection{Robustness}
\begin{figure}[h]
  \centering
  \includegraphics[width=3.3in]{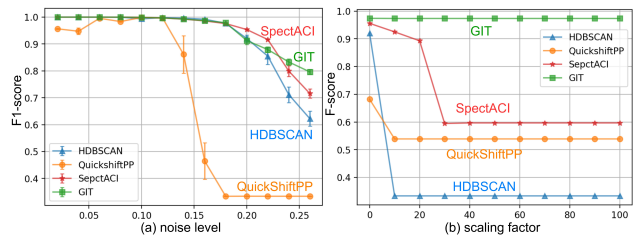}
  \caption{Performance comparison under different noise levels and scales. In (a), we change the (Gaussian) noise on Moons from 0.02 to 0.26 with step 0.02, and plot the average F1-score and its variance over 10 random seeds. In (b),  we show the F1-score for each scaling factor ranging from 1 to 100 on the mixed Circles.}
  \label{fig:robustness_change}
\end{figure}

\paragraph{Objective and Setting.} To study the robustness to shapes, noises and scales of various algorithms, we: 1) firstly compare all algorithms on synthetic datasets with complex shapes, such as Circles and Moons, and show the top-3 (F1-score) results, 2) secondly compare GIT with top-3 competitors on Moons (with Gaussian noise) and Impossible (with Uniform noises), and 3) thirdly evaluate these methods under the mixed multi-scale data, e.g. a new dataset containing original cricles and enlarged circles in Fig.~\ref{tab:noises_scales}. We report F1-scores on different noise or scaling levels in Fig.~\ref{fig:robustness_change}.

\begin{table}[H]
  \small
  \centering
  \resizebox{\columnwidth}{!}{
  \begin{tabular}{ c | ccc }
    \toprule
    Data & Top1 & Top2 & Top3\\ 
    \hline

    \begin{minipage}[b]{0.25\columnwidth}
		\centering
		\raisebox{-.5\height}{\includegraphics[width=\linewidth]{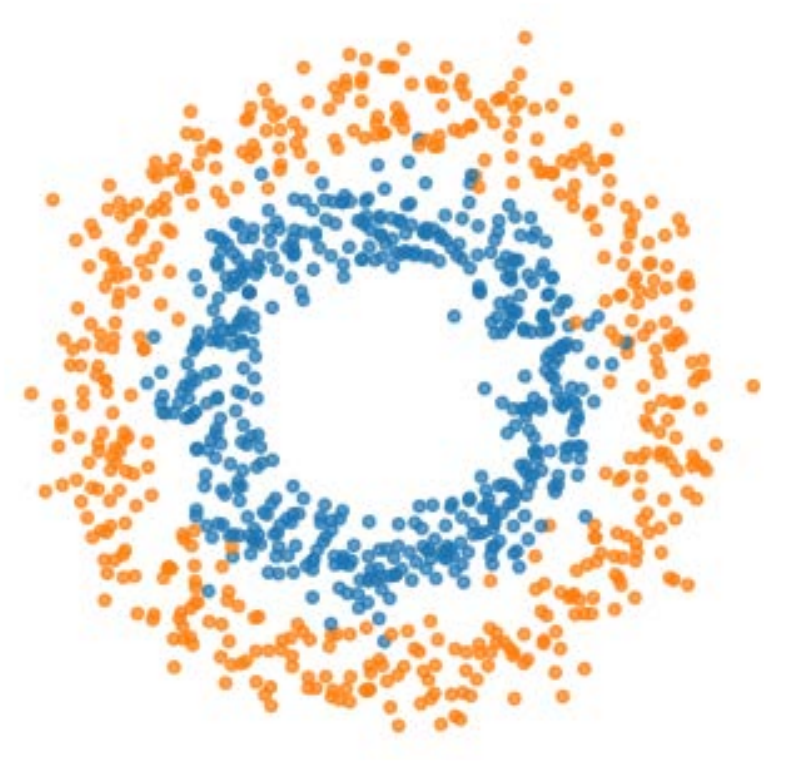}}
		\vspace*{-4mm}
		\caption*{Circles}
	\end{minipage}
    & \begin{minipage}[b]{0.25\columnwidth}
		\centering
		\raisebox{-.5\height}{\includegraphics[width=\linewidth]{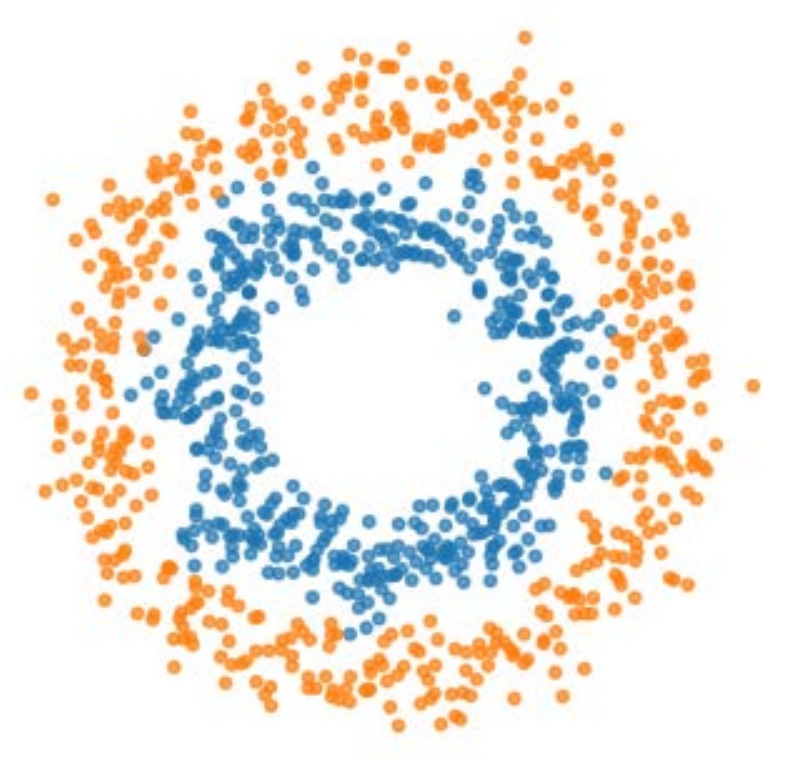}}
		\vspace*{-4mm}
		\caption*{\textbf{GIT}}
	\end{minipage}
    & \begin{minipage}[b]{0.25\columnwidth}
		\centering
		\raisebox{-.5\height}{\includegraphics[width=\linewidth]{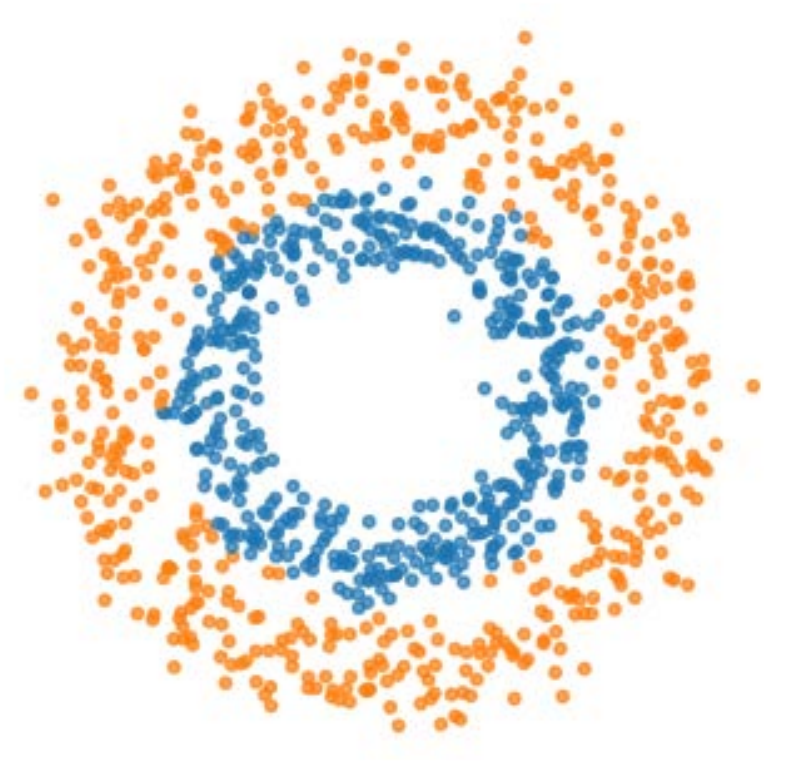}}
		\vspace*{-4mm}
		\caption*{SA}
	\end{minipage}
	& \begin{minipage}[b]{0.25\columnwidth}
		\centering
		\raisebox{-.5\height}{\includegraphics[width=\linewidth]{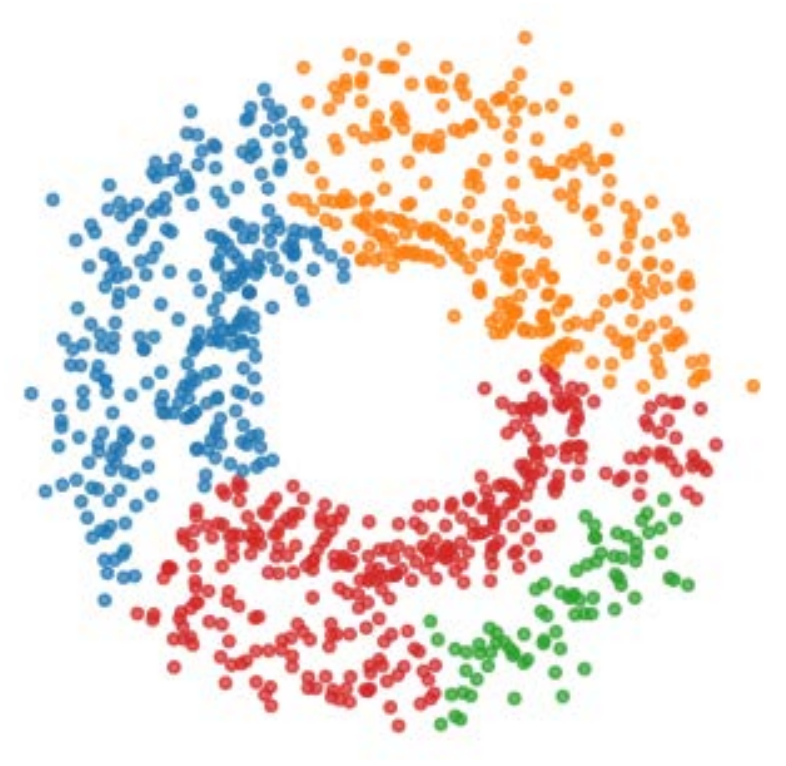}}
		\vspace*{-4mm}
		\caption*{QSP}
	\end{minipage}\\

    \begin{minipage}[b]{0.25\columnwidth}
		\centering
		\raisebox{-.5\height}{\includegraphics[width=\linewidth]{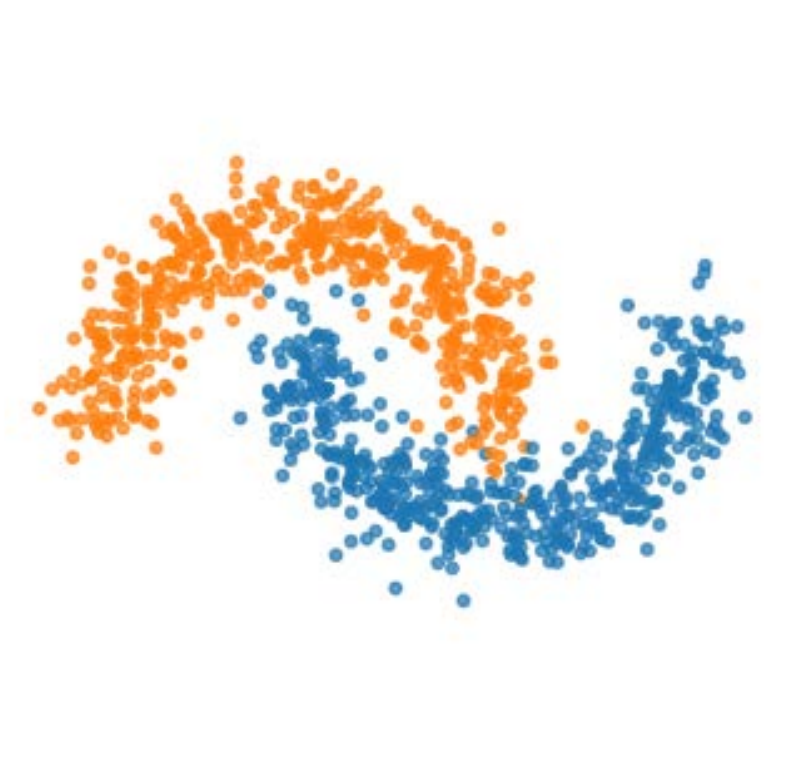}}
		\vspace*{-4mm}
		\caption*{Moons}
	\end{minipage}
    & \begin{minipage}[b]{0.25\columnwidth}
		\centering
		\raisebox{-.5\height}{\includegraphics[width=\linewidth]{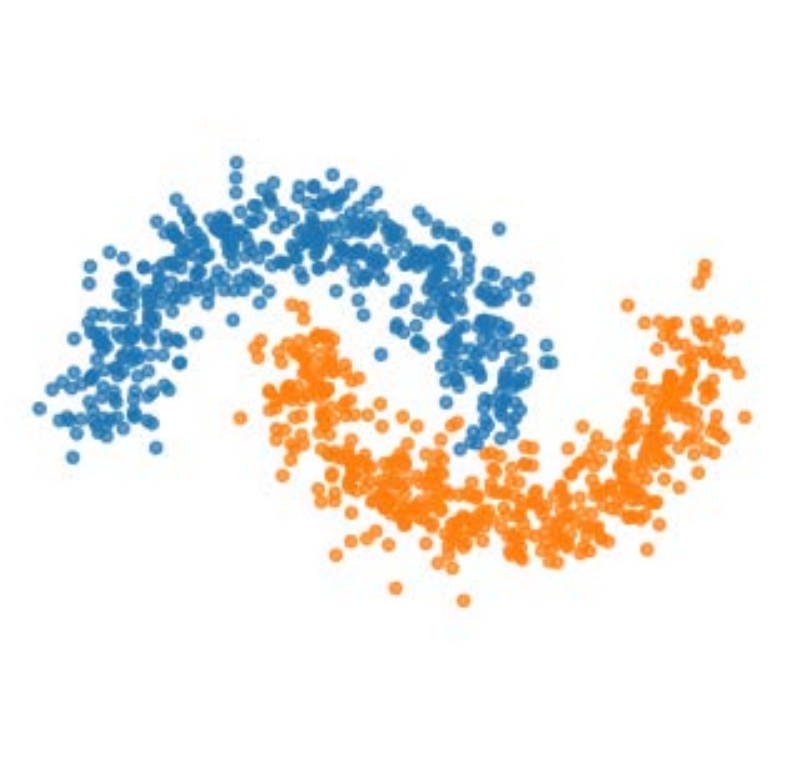}}
		\vspace*{-4mm}
		\caption*{\textbf{GIT}, QSP, SA}
	\end{minipage}
    & \begin{minipage}[b]{0.25\columnwidth}
		\centering
		\raisebox{-.5\height}{\includegraphics[width=\linewidth]{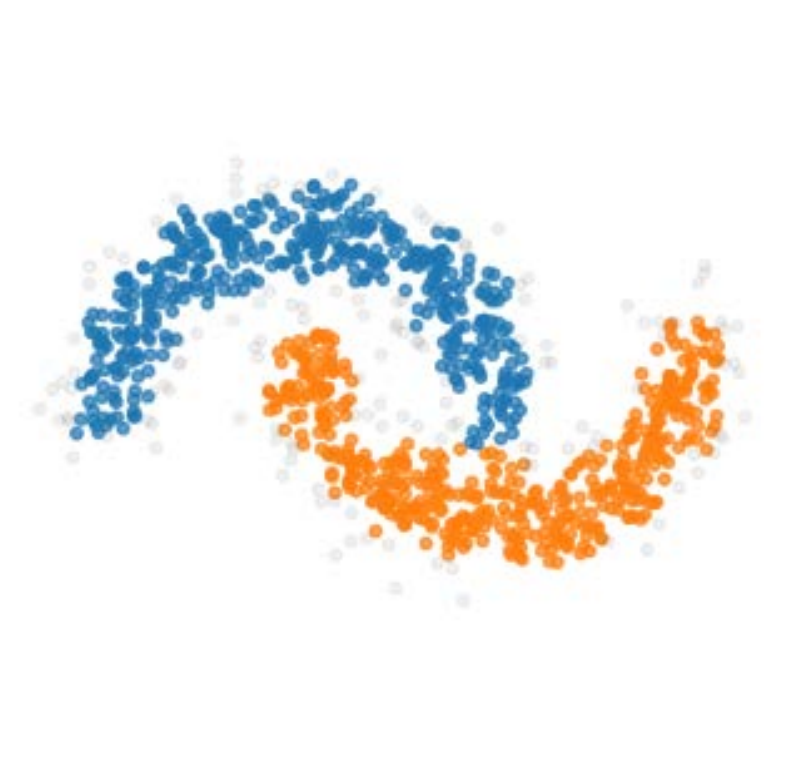}}
		\vspace*{-4mm}
		\caption*{HDB}
	\end{minipage}
	& \begin{minipage}[b]{0.25\columnwidth}
		\centering
		\raisebox{-.5\height}{\includegraphics[width=\linewidth]{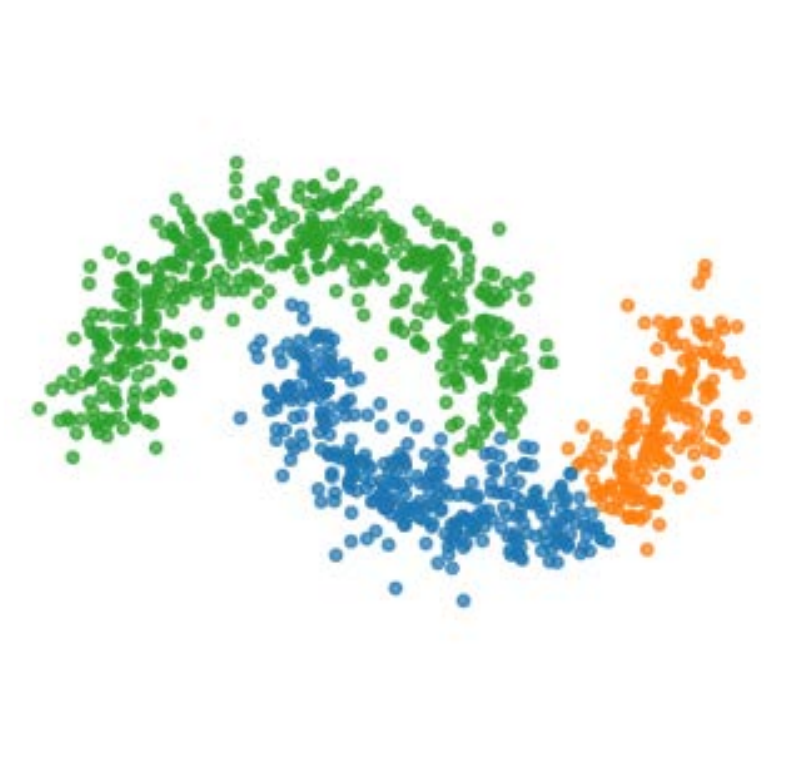}}
		\vspace*{-4mm}
		\caption*{FSF}
	\end{minipage}\\

    \begin{minipage}[b]{0.25\columnwidth}
		\centering
		\raisebox{-.5\height}{\includegraphics[width=\linewidth]{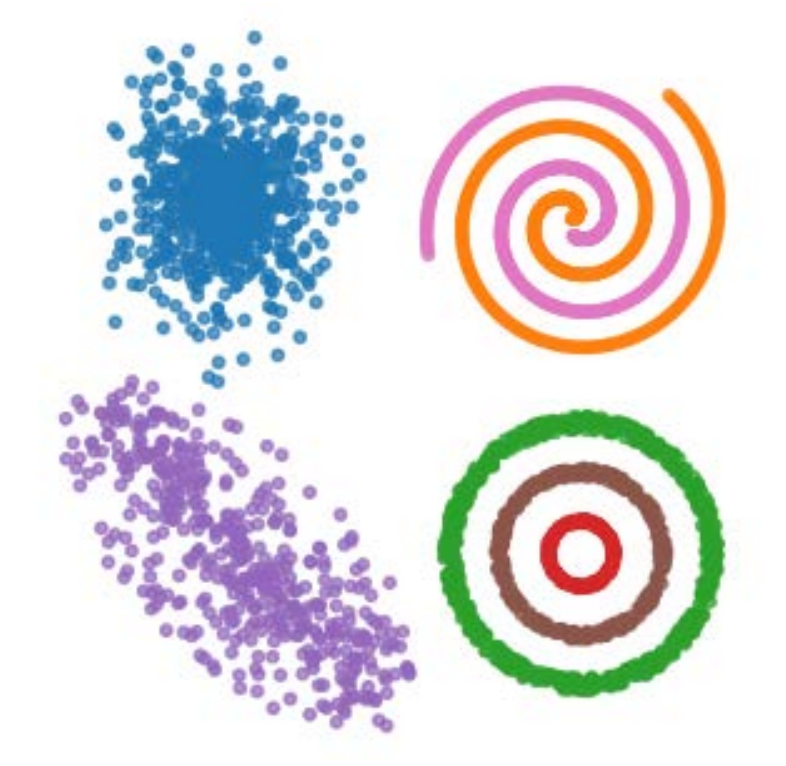}}
		\vspace*{-4mm}
		\caption*{Impossible}
	\end{minipage}
    & \begin{minipage}[b]{0.25\columnwidth}
		\centering
		\raisebox{-.5\height}{\includegraphics[width=\linewidth]{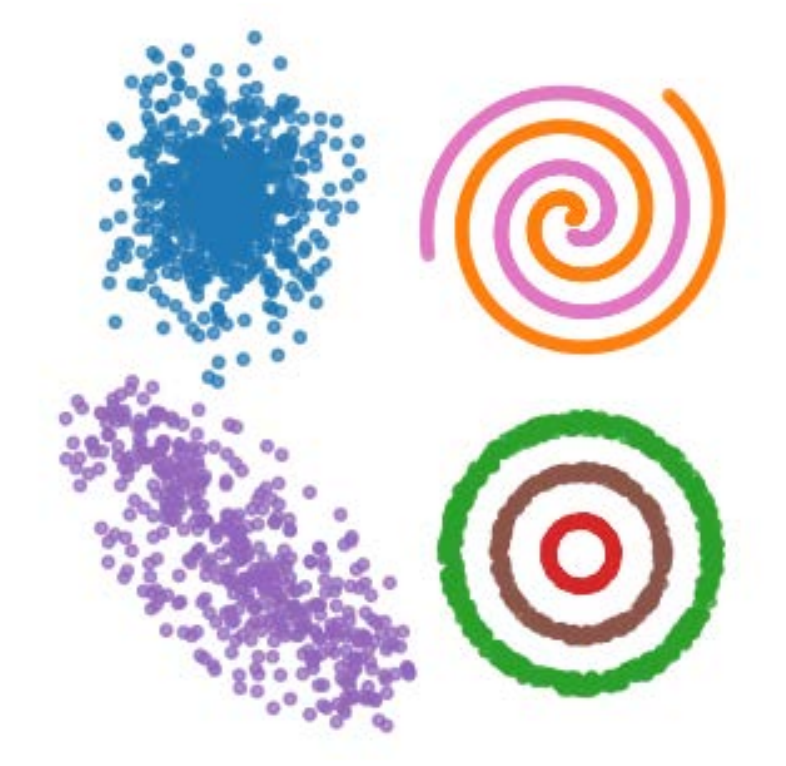}}
		\vspace*{-4mm}
		\caption*{\textbf{GIT}}
	\end{minipage}
    & \begin{minipage}[b]{0.25\columnwidth}
		\centering
		\raisebox{-.5\height}{\includegraphics[width=\linewidth]{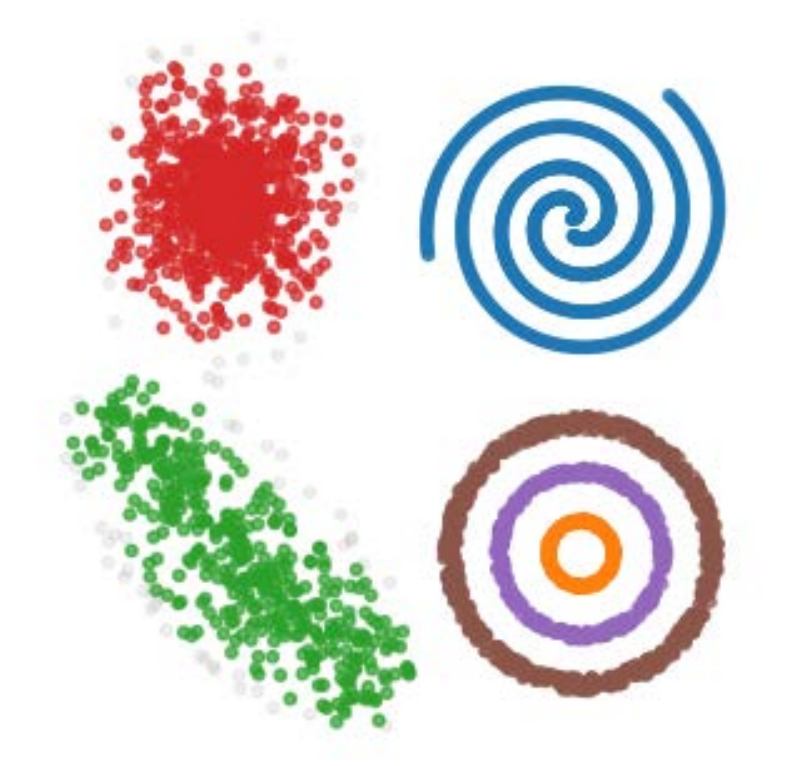}}
		\vspace*{-4mm}
		\caption*{HDB}
	\end{minipage}
	& \begin{minipage}[b]{0.25\columnwidth}
		\centering
		\raisebox{-.5\height}{\includegraphics[width=\linewidth]{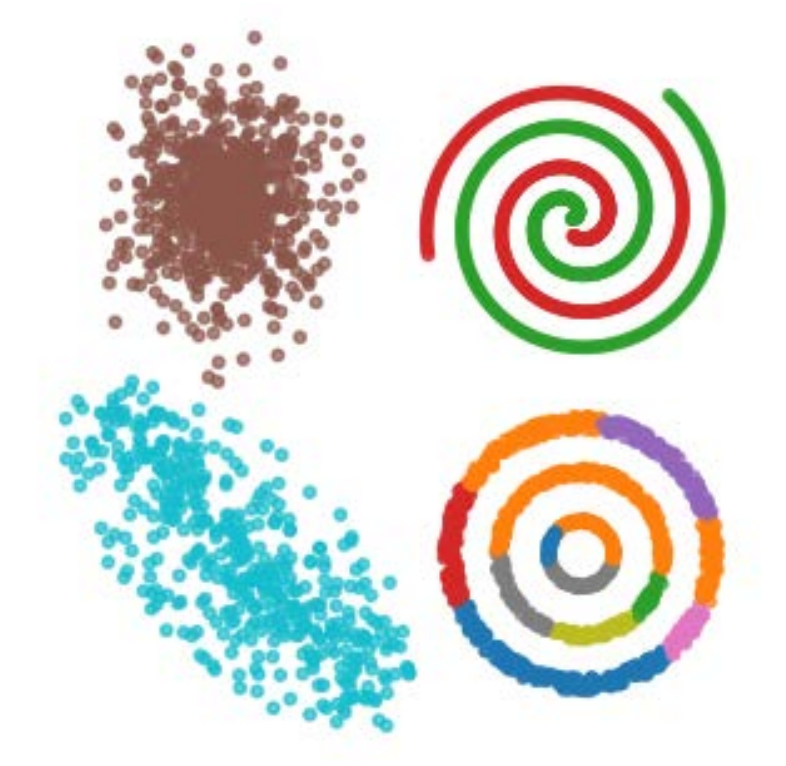}}
		\vspace*{-4mm}
		\caption*{QSP}
	\end{minipage}\\

    \begin{minipage}[b]{0.25\columnwidth}
		\centering
		\raisebox{-.5\height}{\includegraphics[width=\linewidth]{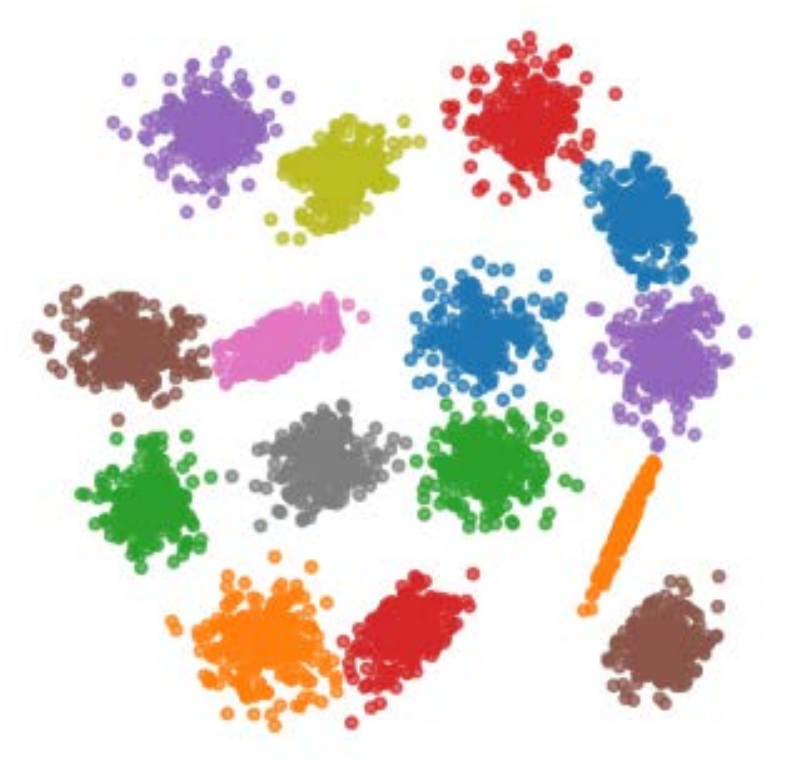}}
		\vspace*{-4mm}
		\caption*{S-sets}
	\end{minipage}
    & \begin{minipage}[b]{0.25\columnwidth}
		\centering
		\raisebox{-.5\height}{\includegraphics[width=\linewidth]{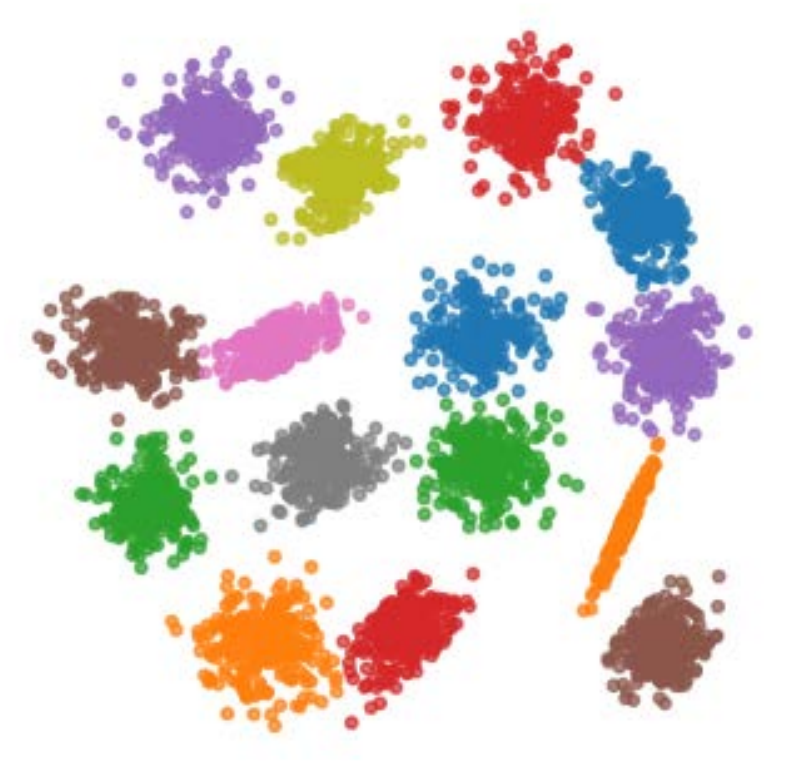}}
		\vspace*{-4mm}
		\caption*{\scriptsize \textbf{GIT}, FSF, QSP, KM}
	\end{minipage}
    & \begin{minipage}[b]{0.25\columnwidth}
		\centering
		\raisebox{-.5\height}{\includegraphics[width=\linewidth]{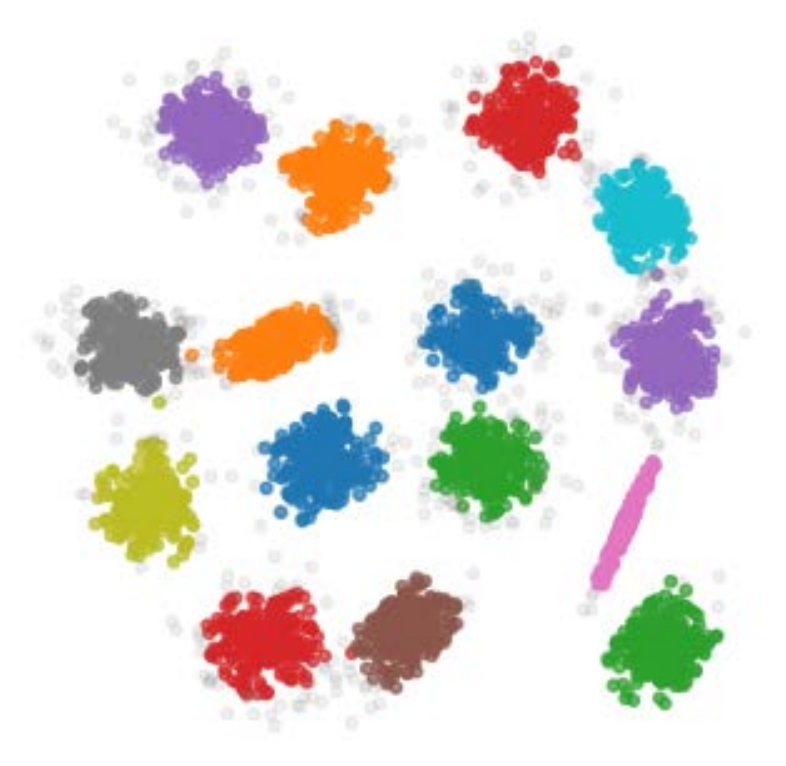}}
		\vspace*{-4mm}
		\caption*{DPA, HDB}
	\end{minipage}
	& \begin{minipage}[b]{0.25\columnwidth}
		\centering
		\raisebox{-.5\height}{\includegraphics[width=\linewidth]{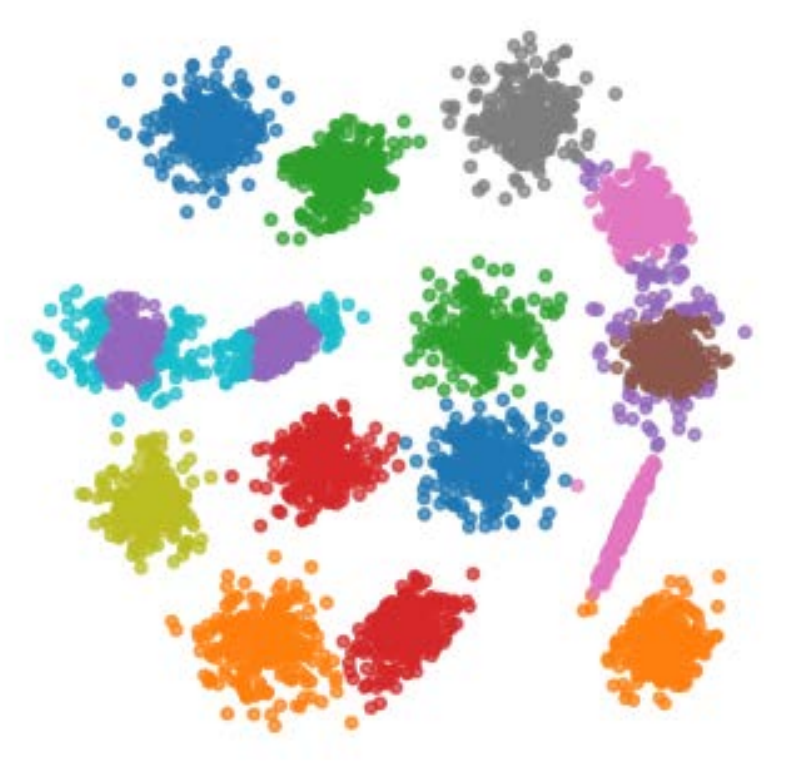}}
		\vspace*{-4mm}
		\caption*{SA}
	\end{minipage}\\

    \begin{minipage}[b]{0.25\columnwidth}
		\centering
		\raisebox{-.5\height}{\includegraphics[width=\linewidth]{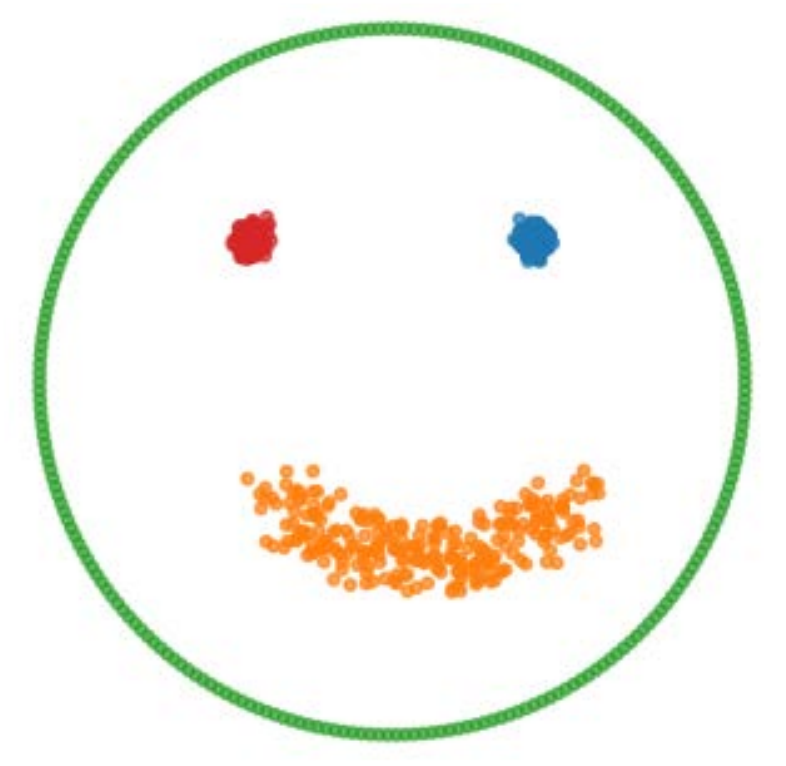}}
		\vspace*{-4mm}
		\caption*{Smiles}
	\end{minipage}
    & \begin{minipage}[b]{0.25\columnwidth}
		\centering
		\raisebox{-.5\height}{\includegraphics[width=\linewidth]{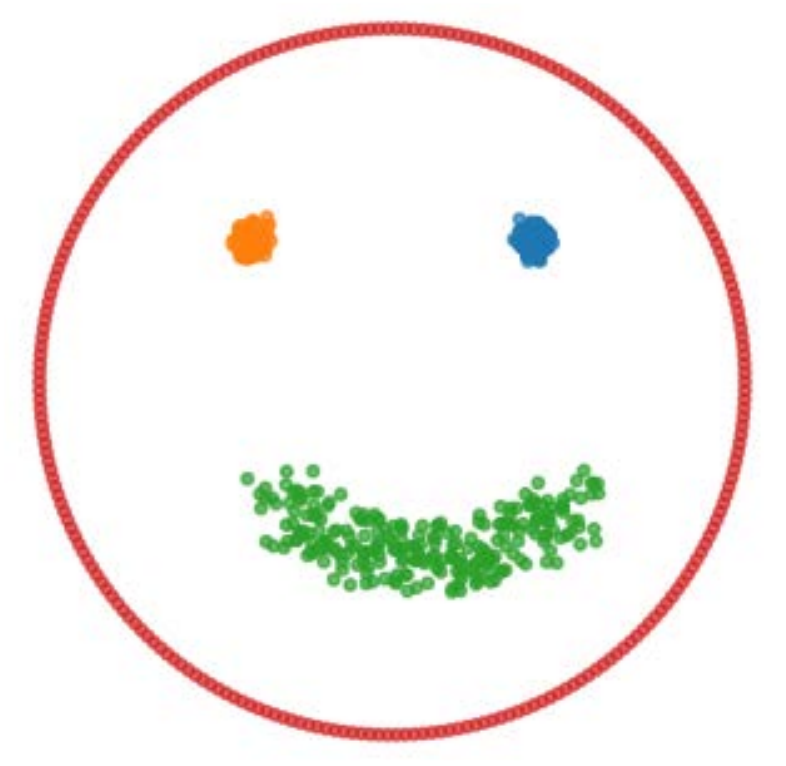}}
		\vspace*{-4mm}
		\caption*{ \scriptsize \textbf{GIT}, HDB, SA, SC}
	\end{minipage}
    & \begin{minipage}[b]{0.25\columnwidth}
		\centering
		\raisebox{-.5\height}{\includegraphics[width=\linewidth]{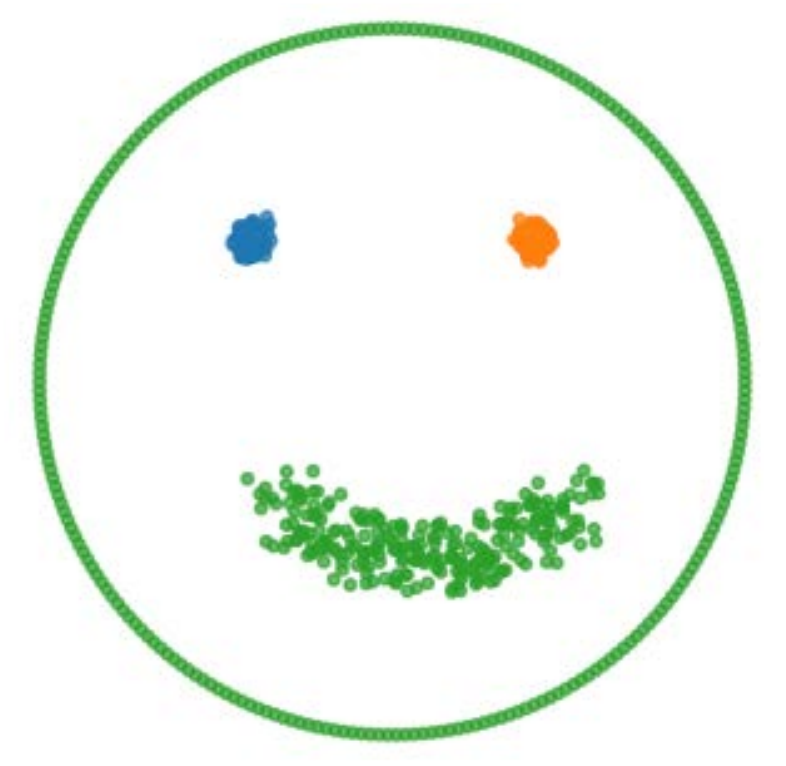}}
		\vspace*{-4mm}
		\caption*{DPA}
	\end{minipage}
	& \begin{minipage}[b]{0.25\columnwidth}
		\centering
		\raisebox{-.5\height}{\includegraphics[width=\linewidth]{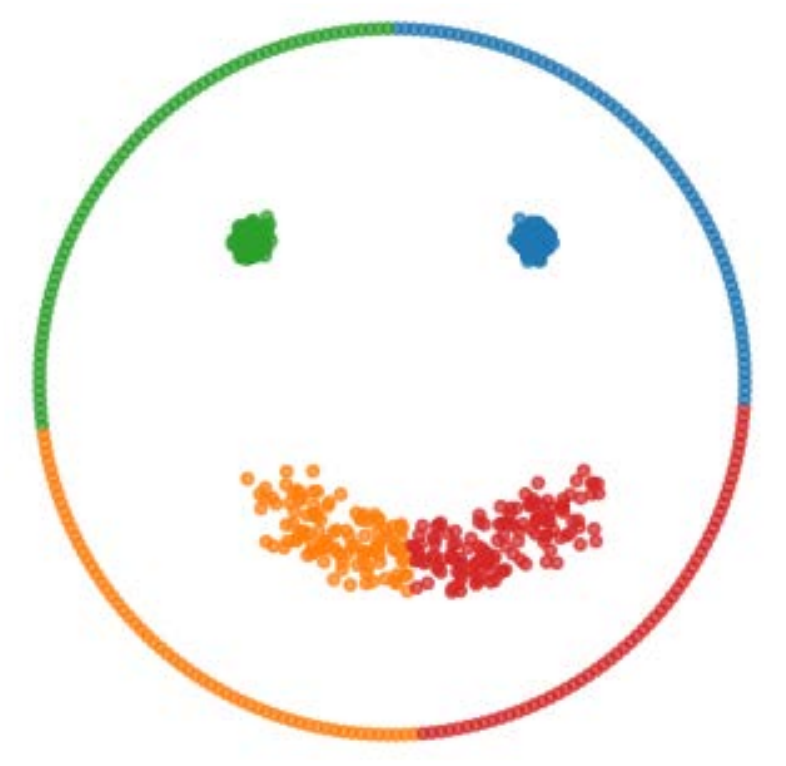}}
		\vspace*{-4mm}
		\caption*{KM}
	\end{minipage}\\ 
  \bottomrule
  \end{tabular}
  }
  \caption{Top-3 (F1-score) results on synthetic datasets with complex shapes. The first column is the ground truth data. Other columns show the clustering results from top-1 to top-3. We mark the corresponding method abbreviation under each result. }
  \label{tab:shapes}
\end{table}

\paragraph{Result and Analysis.} For shapes, we evaluate all baselines but only visualize the top-3 results in Table.~\ref{tab:shapes}. GIT is the best one to get the resonable clusters on all synthetic datasets and sup-optimum methods are SA, QSP, HDB and SC in turns. As expected, many baseline methods cannot deal with complex distributions (e.g., the Impossible dataset) for the lack of ability to identify global data structures, while GIT handles it well. For noises and scales, we compare GIT with SA and QSP in Table.~\ref{tab:noises_scales}, from which we find that GIT is more robust against noise and data scales, whereas SA and QSP both fail. What's more, we plot the F1-scores of GIT, SA, QSP and HDB under different noise levels and scaling factors (Fig.~\ref{fig:robustness_change}), where GIT achieve the highest F1-score with smallest variance on the extreme situations. We believe the robustness comes from two aspects:  On the one hand, the well-designed intensity function helps handle the multi-scale issue because the kNN is invariant to scales, as proved in the Appendix (Theorem 1). On the other hand, the local and global clustering mechanism is helpful to anti-noise. Firstly, the process of local clustering detection is reliable because noises usually have a limited effect on the relative point intensities. Secondly, connectivity strength between two local clusters is also robust because it depends on all the adjacent boundary points, not just a few noisy points. As both nodes and edges of the topo-graph are robust to noise, the clustering results are undoubtedly robust.

\begin{table}[h]
  \centering
  \resizebox{\columnwidth}{!}{
  \begin{tabular}{ c | ccc }
        \toprule
         Data & \textbf{GIT} & SA & QSP\\ 
        
        \hline
        
        \begin{minipage}[b]{0.25\columnwidth}
    		\centering
    		\raisebox{-.5\height}{\includegraphics[width=\linewidth]{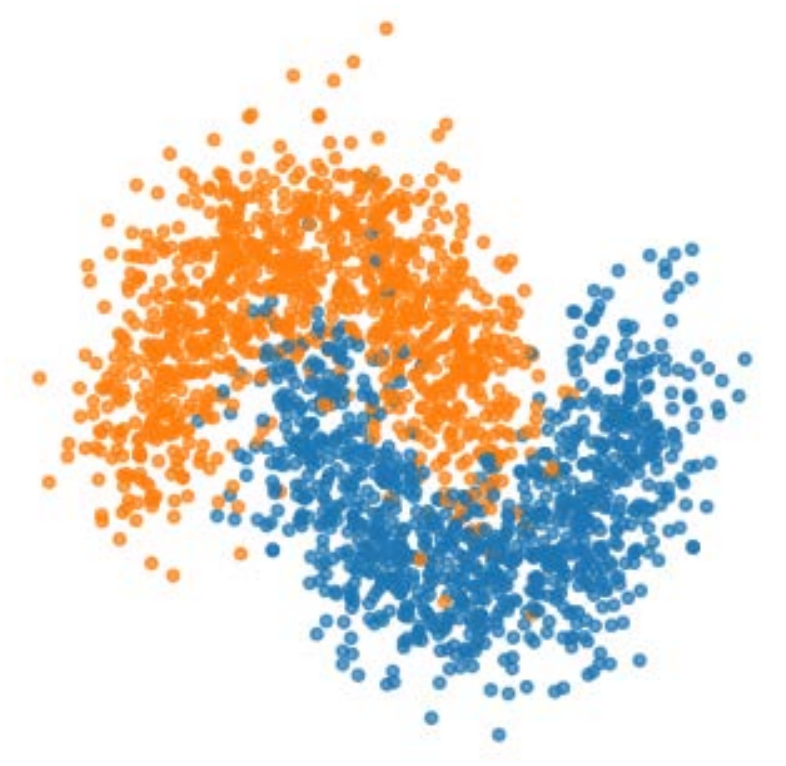}}
    	\end{minipage}
        & \begin{minipage}[b]{0.25\columnwidth}
    		\centering
    		\raisebox{-.5\height}{\includegraphics[width=\linewidth]{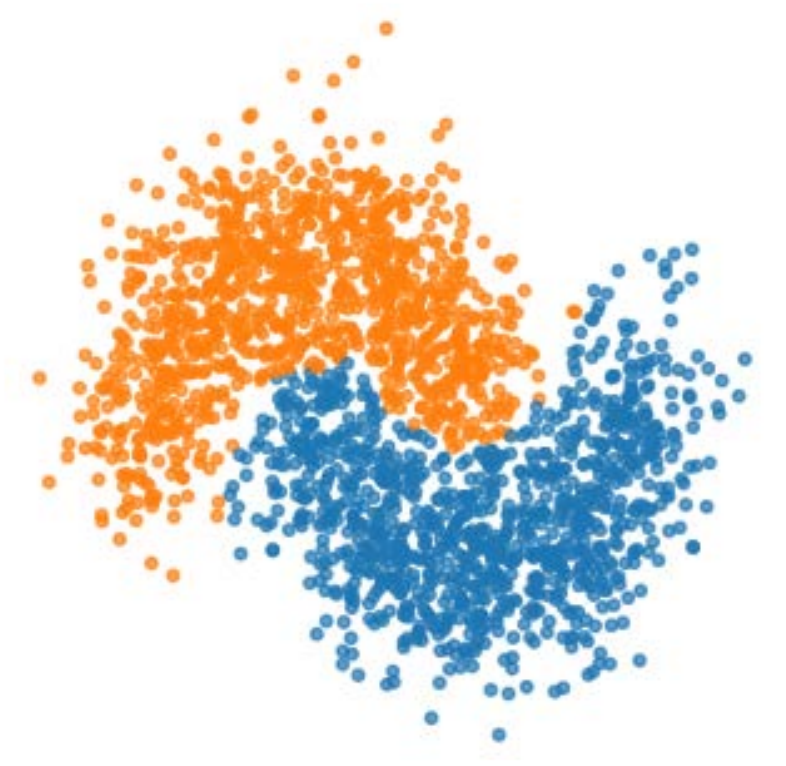}}
    	\end{minipage}
        & \begin{minipage}[b]{0.25\columnwidth}
    		\centering
    		\raisebox{-.5\height}{\includegraphics[width=\linewidth]{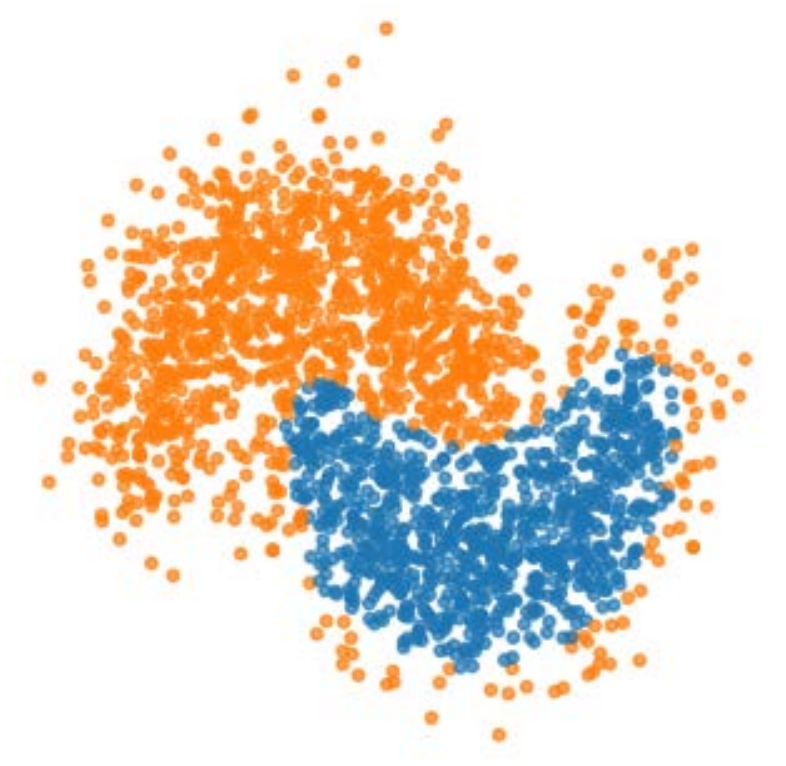}}
    	\end{minipage}
    	& \begin{minipage}[b]{0.25\columnwidth}
    		\centering
    		\raisebox{-.5\height}{\includegraphics[width=\linewidth]{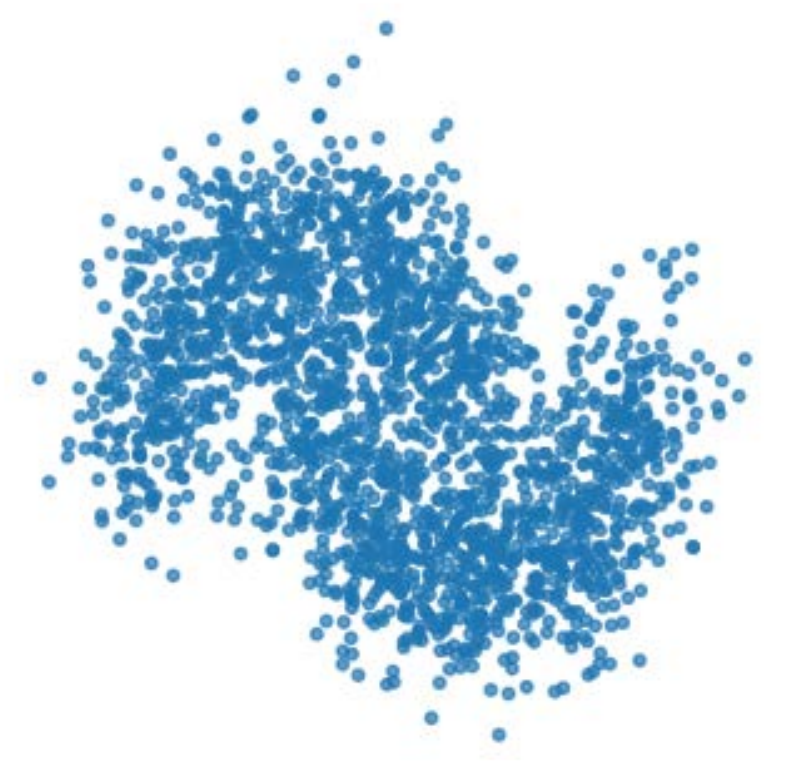}}
    	\end{minipage}\\

    	\begin{minipage}[b]{0.25\columnwidth}
    		\centering
    		\raisebox{-.5\height}{\includegraphics[width=\linewidth]{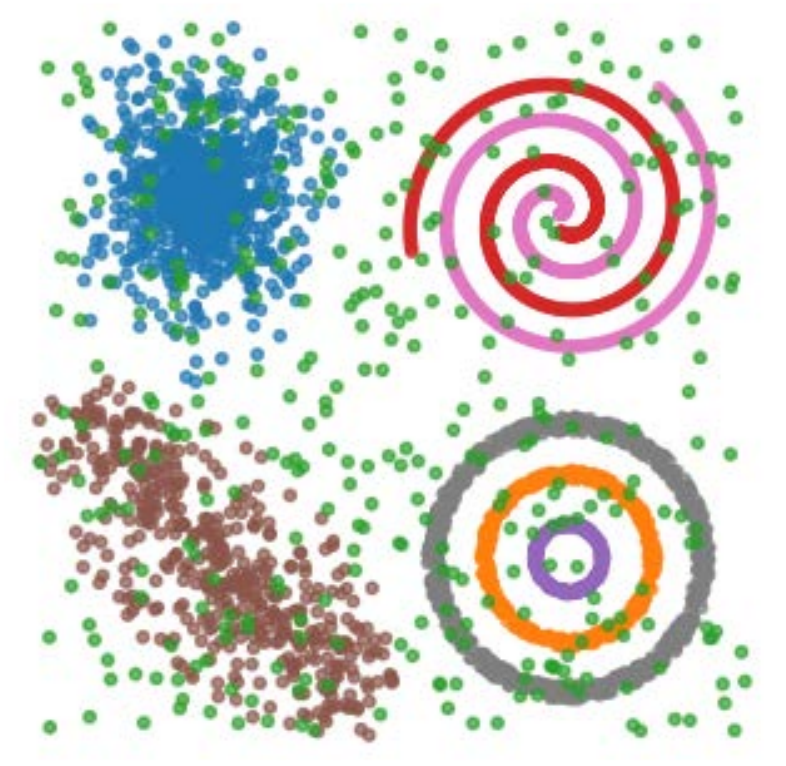}}
    	\end{minipage}
        & \begin{minipage}[b]{0.25\columnwidth}
    		\centering
    		\raisebox{-.5\height}{\includegraphics[width=\linewidth]{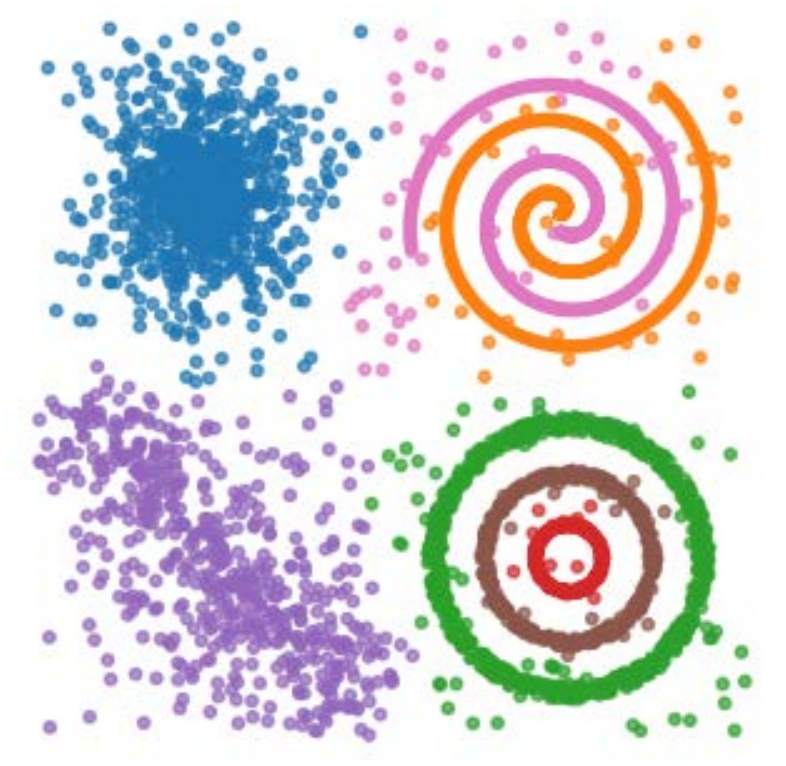}}
    	\end{minipage}
        & \begin{minipage}[b]{0.25\columnwidth}
    		\centering
    		\raisebox{-.5\height}{\includegraphics[width=\linewidth]{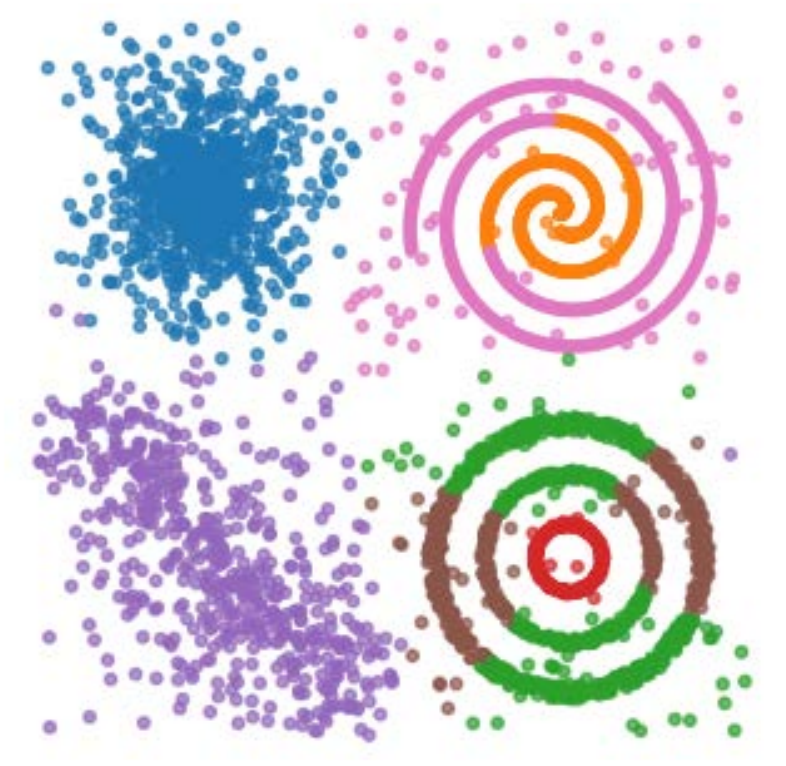}}
    	\end{minipage}
    	& \begin{minipage}[b]{0.25\columnwidth}
    		\centering
    		\raisebox{-.5\height}{\includegraphics[width=\linewidth]{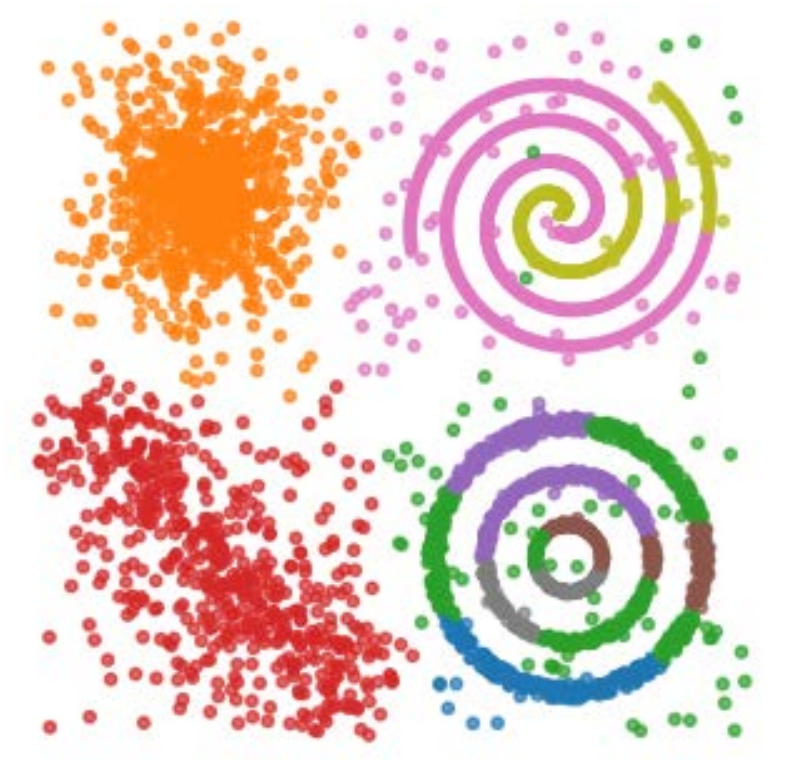}}
    	\end{minipage}\\

    	\begin{minipage}[b]{0.32\columnwidth}
    		\centering
    		\raisebox{-.5\height}{\includegraphics[width=\linewidth]{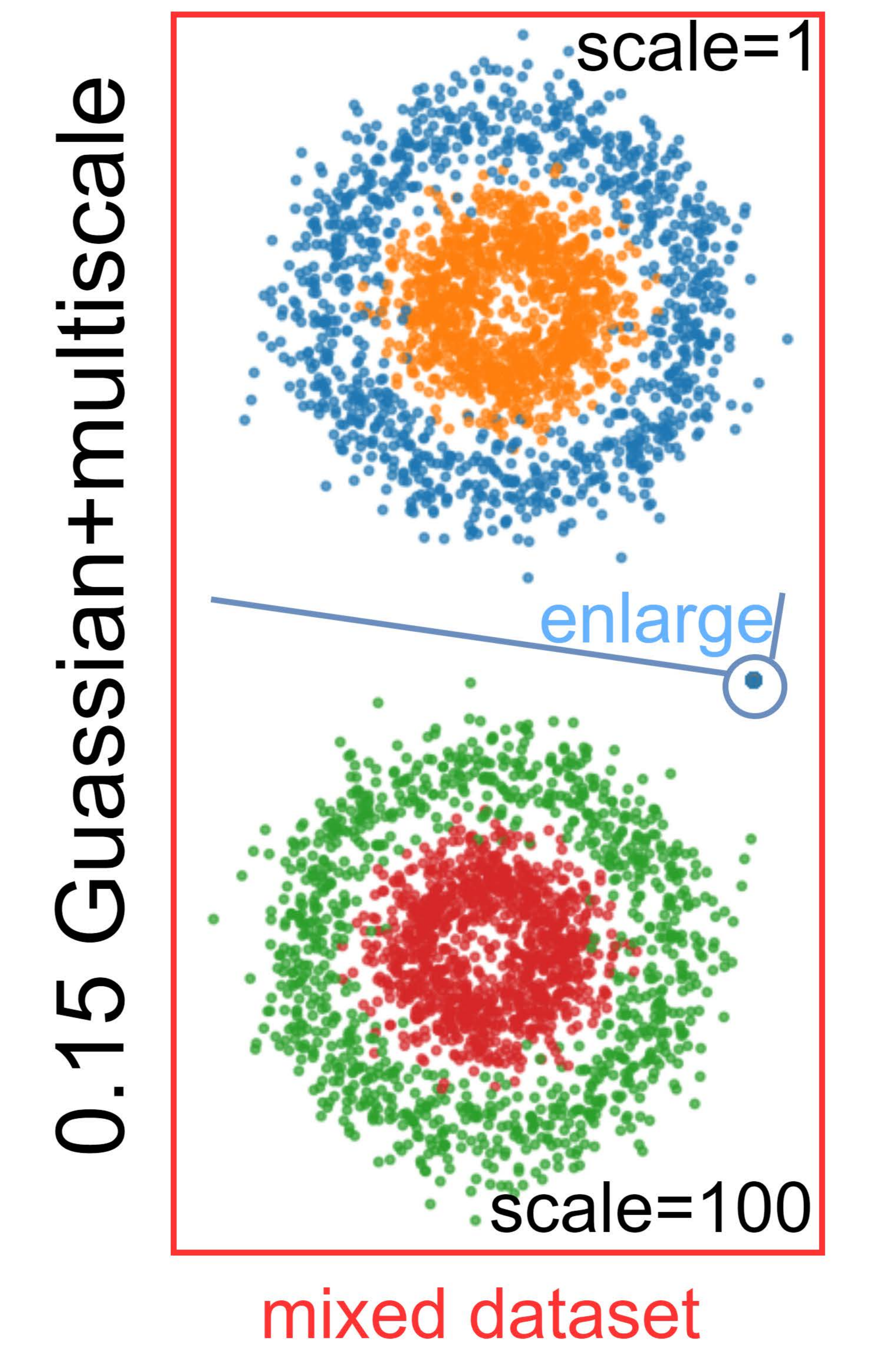}}
    	\end{minipage}
        & \begin{minipage}[b]{0.25\columnwidth}
    		\centering
    		\raisebox{-.5\height}{\includegraphics[width=\linewidth]{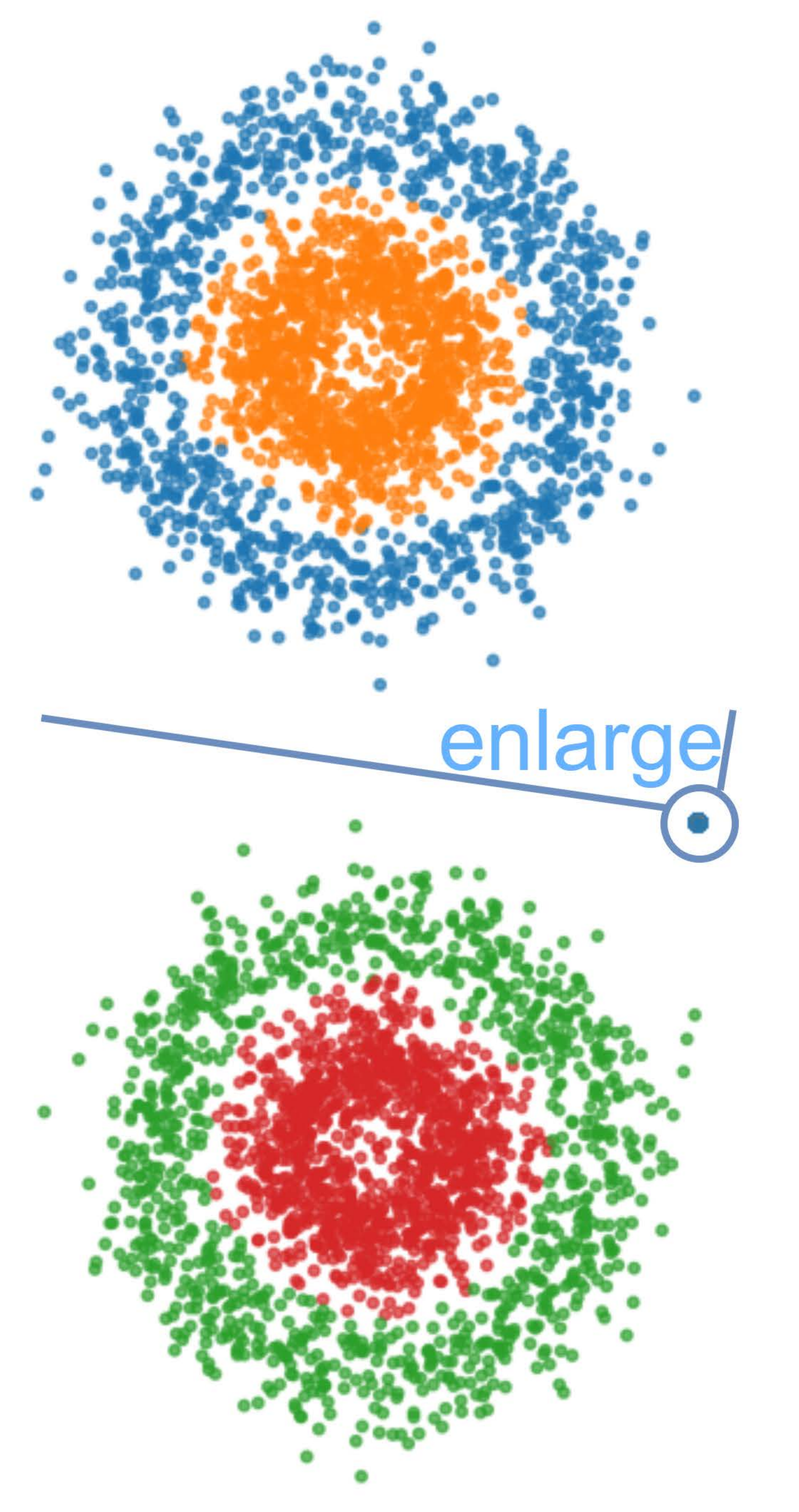}}
    	\end{minipage}
        & \begin{minipage}[b]{0.25\columnwidth}
    		\centering
    		\raisebox{-.5\height}{\includegraphics[width=\linewidth]{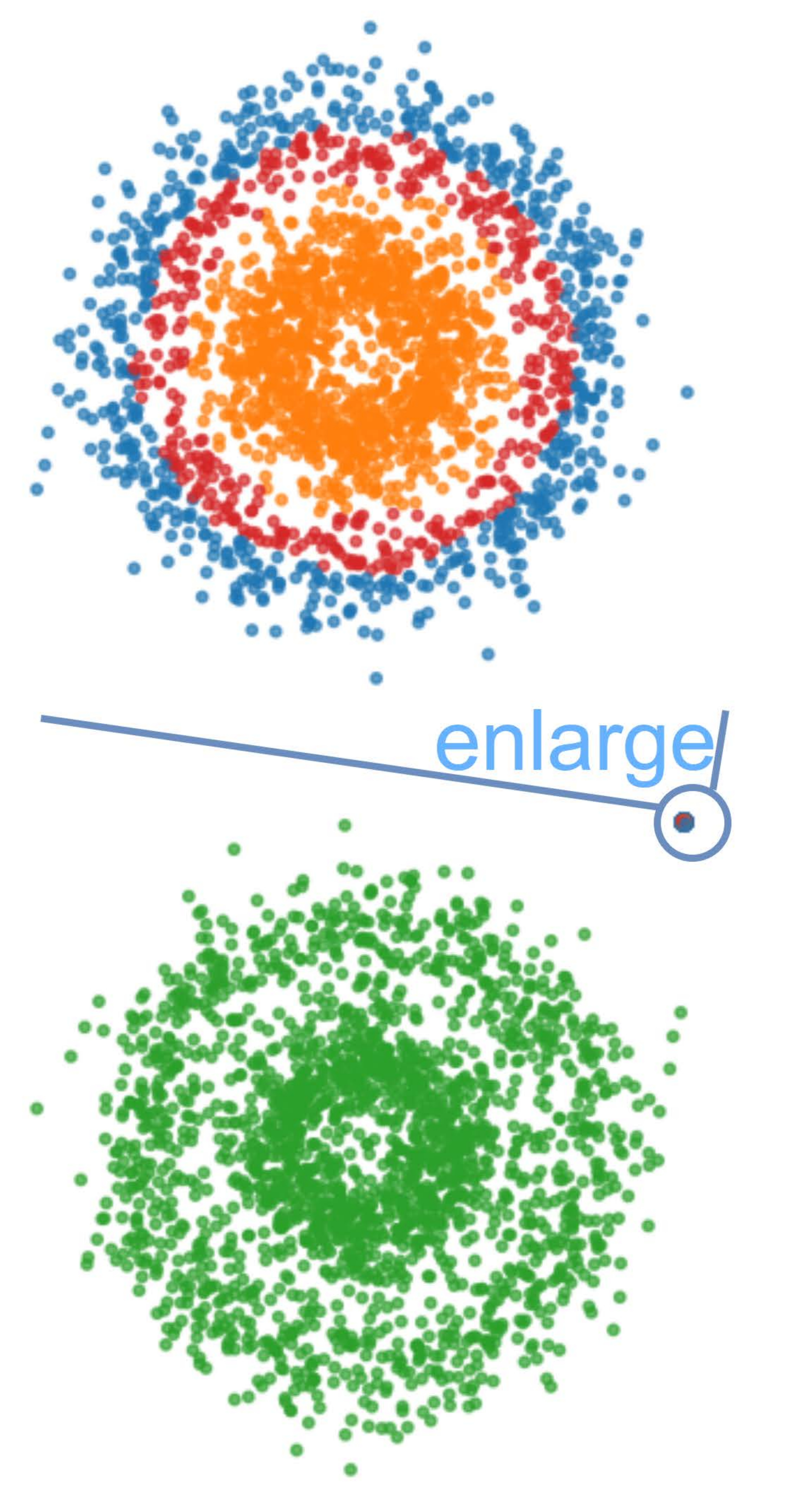}}
    	\end{minipage}
    	& \begin{minipage}[b]{0.25\columnwidth}
    		\centering
    		\raisebox{-.5\height}{\includegraphics[width=\linewidth]{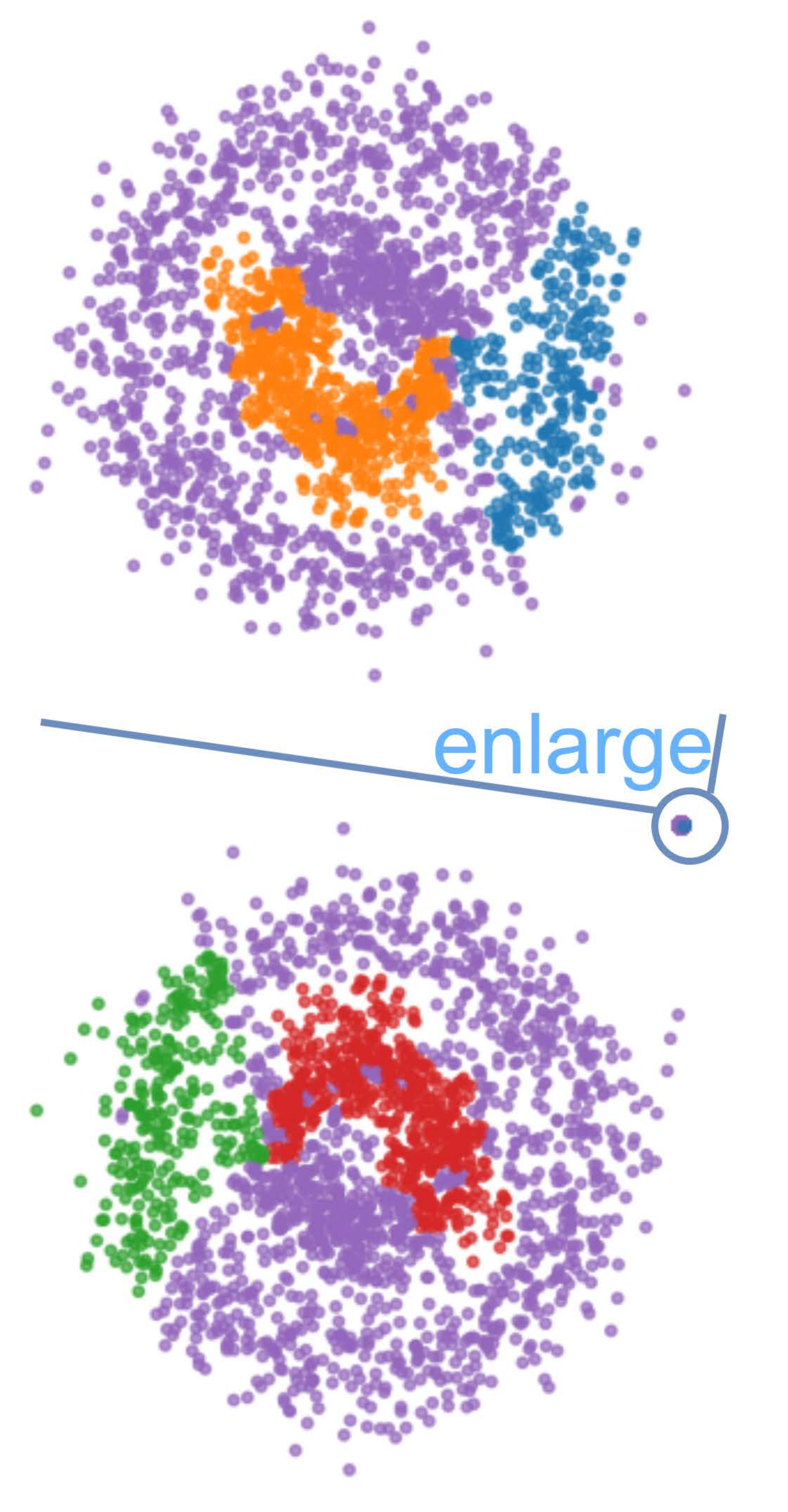}}
    	\end{minipage}\\ 
    	
    	\bottomrule
  \end{tabular}
  }
  \caption{Comparing GIT with SA and QSP on noisy and multi-scale datasets. The first column is the raw data and others are results of GIT, SA and QSP. We add 0.2 Gaussian noise on Circles, 0.1 Uniform noise on Impossible and mix two Circles as the multi-scale data, where one Circles is 100 times larger than the other.}
  \label{tab:noises_scales}
\end{table}

\vspace{-5mm}

\subsection{Dimension Reduction + GIT}
\label{sec:dim_reduction_GIT}
\paragraph{Objective and Setting.} To alleviate the problem of dimension curse, dimension reduction methods are often used together with clustering algorithms, such as PCA and Autoencoder. We study how far GIT outperforms competitors under these settings. As to PCA, we directly use scikit-learn's API to project raw data into $h$-dimensional space. For the autoencoder, we construct MLP using PyTorch with the following structure: 784-512-256-128-$h$ (enc) and $h$-128-256-512-784 (dec). We use Adam to optimize the autoencoder up to 100 epochs with the learning rate 0.001. In both of these settings, we project 60k samples (training set, dim=784) to $h$-dimensional space, where $h\in \{5,10,20,30\}$. Finally, we use the same embeddings as the inputs of various clustering methods and report the F1-score.

\begin{figure}[h]
    \centering
    \includegraphics[width=3.3in]{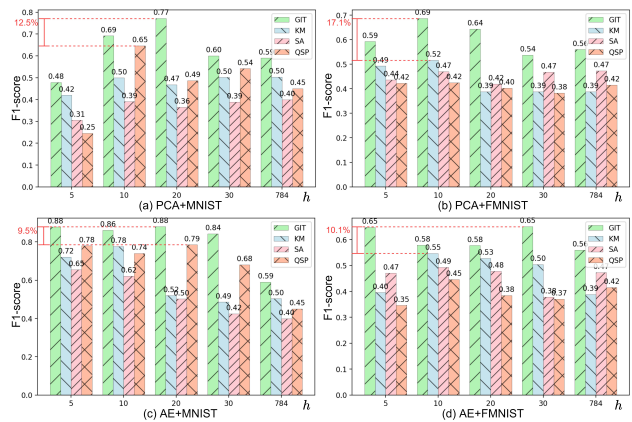}
    \caption{Dimension reduction + Clustering. The x-axis and y-axis are the projected dimension $h$ and F1-score. GIT outperforms competitors by up to 12.5\% (on PCA+MNIST), 17.1\% (on PCA+FMNIST), 9.5\% (on AE+MNSIT) and 10.4\% (on AE+FMNIST).}
    \label{fig:dim_reduction}
\end{figure}

\begin{figure}[h]
  \centering
  \includegraphics[width=3.3in]{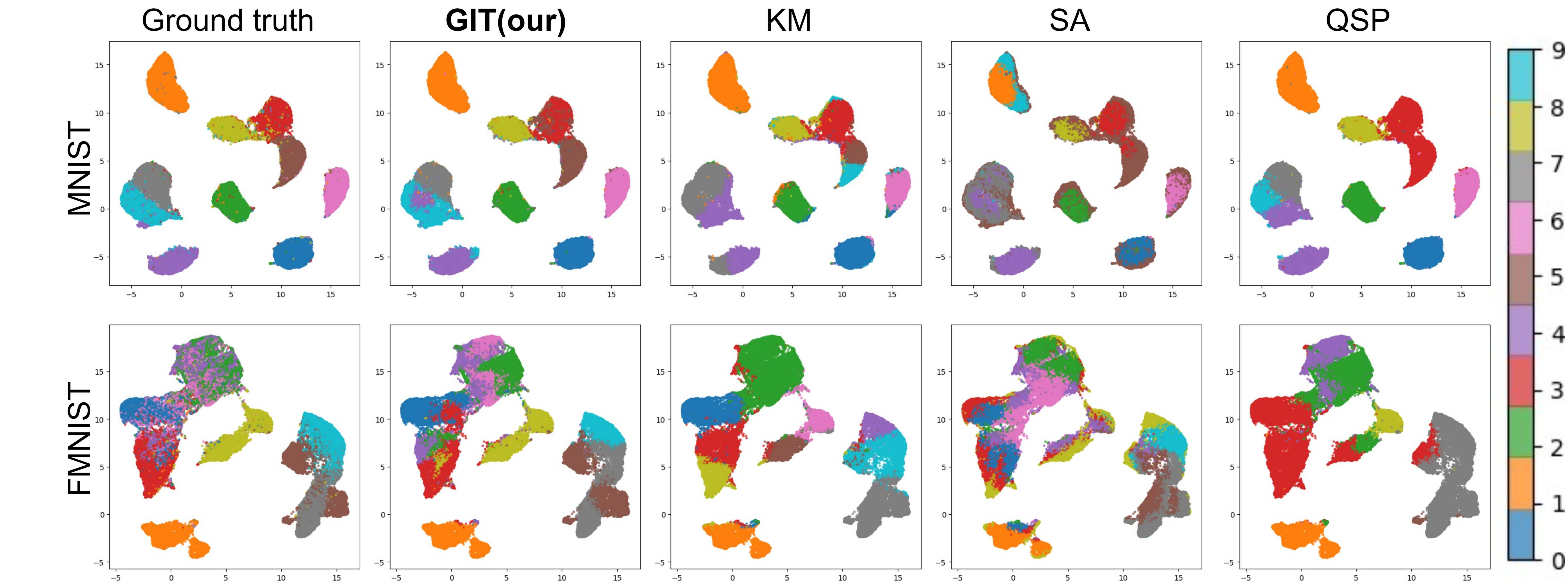}
  \caption{Visualization of results ($h=5$) on MNIST and FMNIST with UMAP. GIT provide more accurate clusters than competitors. }
  \label{fig:visualization}
\end{figure}

\paragraph{Result and Analysis.} As shown in Fig.~\ref{fig:dim_reduction}, both PCA and AE bring consistent improvement of F1-score for clustering methods excluding SA. It is worth pointing out that GIT consistently outperforms competitors on all settings, as shown in Fig.~\ref{fig:dim_reduction}. We further visualize the results of AE+MNSIT and AE+FMNIST ($h$=5) via UMAP \cite{mcinnes2018umap}, and Fig.~\ref{fig:visualization} shows that GIT generates more reasonable clusters than other algorithms.

\vspace{-5mm}

\vspace{-2mm}
\section{Conclusion}
We  propose  a  novel  clustering  algorithm GIT to  achieve  better \textbf{A}ccuracy, \textbf{R}obustness to noises and scales, \textbf{I}nterpretability with accaptable \textbf{S}peed and \textbf{E}asy to use (ARISE), considering both local and global data structures. Compared with previous works, the proper usage of global structure is the key to GIT's accuracy gain. Both the intensity-based local cluster detection and well-designed topo-graph connectivity make it robust. We believe that GIT will promote the development of cluster analysis in various scientific fields.

\newpage
\newpage
\bibliography{references}

\clearpage
\pdfoutput=1
\section{Appendix}
\subsection{Itensify function}
\label{sup:idensity_function}

\paragraph{Definition 1.} 
If $|\mathcal{N}_{\boldsymbol{x}}|=|\mathcal{N}_{\boldsymbol{y}}|=k$ and points in $\mathcal{N}_{\boldsymbol{x}}=\{\boldsymbol{x}_i\}_{i=1}^{k}, \mathcal{N}_{\boldsymbol{y}}=\{\boldsymbol{y}_i\}_{i=1}^{k}$ can be re-indexed to form sequences $(\boldsymbol{x}_1,\boldsymbol{x}_2,\dots,\boldsymbol{x}_k)$ and $(\boldsymbol{y}_1,\boldsymbol{y}_2,\dots,\boldsymbol{y}_k)$, satisfying $\sup_{1\leq i \leq k}|d(\boldsymbol{x}_i,\boldsymbol{y}_i)| \leq d(\boldsymbol{x},\boldsymbol{y})$, we call that $\mathcal{N}_{\boldsymbol{x}_i} \rightarrow \mathcal{N}_{\boldsymbol{x}_j}$.

\paragraph{Theorem 1.} $f(\boldsymbol{x})$ is invariant to the scale of the data.

\paragraph{Proof 1.} Denote $s$ as the transform scale of dataset and $\boldsymbol{x}'=s\boldsymbol{x}, \boldsymbol{y}'=s\boldsymbol{y}$, we know that $\sigma'_j=s\sigma_j$ for all $0<j<s$. According to Eq.\ref{eq:intensity}, 
$d(\boldsymbol{x'},\boldsymbol{y'})=\sqrt{\sum_{j=1}^{s}\frac{(x'_j-y'_j)^2}{{\sigma'}_j^2}}=\sqrt{\sum_{j=1}^{s}\frac{(s \cdot x_j-s \cdot y_j)^2}{(s\sigma'_j)^2}}=\sqrt{\sum_{j=1}^{s}\frac{(x_j-y_j)^2}{\sigma_j^2}}=d(\boldsymbol{x},\boldsymbol{y})$. Thus $d(\boldsymbol{x},\boldsymbol{y})$ is invariant to the data scale $s$ $\Longrightarrow$ $\mathcal{N}_{\boldsymbol{x}}$ is invariant to $s$. Finally, $f(\boldsymbol{x})$ is invariant to $s$.

\paragraph{Theorem 2.} $f(\boldsymbol{x})$ is Lipschitz continuous under the following condition: (a) if $\boldsymbol{x}_i \rightarrow \boldsymbol{x}_j$, then $\mathcal{N}_{\boldsymbol{x}_i} \rightarrow \mathcal{N}_{\boldsymbol{x}_j}$.

\paragraph{Proof 2.} 
Denote $d_{ij}=d(\boldsymbol{x}_i,\boldsymbol{x}_j)$, $d_i(\boldsymbol{y})=d(\boldsymbol{x}_i,\boldsymbol{y})$ and $|\mathcal{N}_{\boldsymbol{x}_i}|=|\mathcal{N}_{\boldsymbol{x}_j}|=k$. The goal is to prove there are exit a constant $c$, such that $|f(\boldsymbol{x}_i)-f(\boldsymbol{x}_j)|/d_{ij} \leq c$. 

\quad

In case 1, $\exists \text{ constant } a>0$, such that $ d_{ij} \geq a$. Because $|f(\boldsymbol{x}_i)-f(\boldsymbol{x}_j)| \in [0,1]$ and $ a \leq d_{ij}$, we know that $0 \leq |f(\boldsymbol{x}_i)-f(\boldsymbol{x}_j)|/d_{ij}\leq 1/a=constant$. 

\quad

In case 2, $\forall a>0$, $ d_{ij} < a$, which means $\boldsymbol{x}_i \rightarrow \boldsymbol{x}_j$. By the condition (a), we have $\mathcal{N}_{\boldsymbol{x}_i} \rightarrow \mathcal{N}_{\boldsymbol{x}_j}$. From Eq.\ref{eq:intensity} (Intensity Function), we derive that $|f(\boldsymbol{x}_i)-f(\boldsymbol{x}_j)|/d_{ij} 
= \frac{1}{k d_{ij} } | \sum_{\boldsymbol{y} \in \mathcal{N}_{\boldsymbol{x}_i}}e^{-d(\boldsymbol{x}_i,\boldsymbol{y})} -\sum_{\boldsymbol{z} \in \mathcal{N}_{\boldsymbol{x}_j}}e^{-d(\boldsymbol{x}_j,\boldsymbol{z})} |$. 

\quad

According to the definition 1, we can find two sequence $(\boldsymbol{y}_1,\boldsymbol{y}_2,\dots,\boldsymbol{y}_k)$ and $(\boldsymbol{z}_1,\boldsymbol{z}_2,\dots,\boldsymbol{z}_k)$, such that $\boldsymbol{y}_t \rightarrow \boldsymbol{z}_t$, and
{\small
\begin{align}
    k|f(\boldsymbol{x}_i)-f(\boldsymbol{x}_j)|
    &=
    | \sum_{\boldsymbol{y} \in \mathcal{N}_{\boldsymbol{x}_i}}e^{-d(\boldsymbol{x}_i,\boldsymbol{y})} -\sum_{\boldsymbol{z} \in \mathcal{N}_{\boldsymbol{x}_j}}e^{-d(\boldsymbol{x}_j,\boldsymbol{z})} |\\
    &= | \sum_{1\leq t \leq k} (e^{-d(\boldsymbol{x}_i,\boldsymbol{y}_t)}- e^{-d(\boldsymbol{x}_j,\boldsymbol{z}_t)})|\\
    & \leq \sum_{1\leq t \leq k} |e^{-d(\boldsymbol{x}_i,\boldsymbol{y}_t)}- e^{-d(\boldsymbol{x}_j,\boldsymbol{z}_t)}| \\
    & < \sum_{1\leq t \leq k} |d(\boldsymbol{x}_j,\boldsymbol{z}_t)-d(\boldsymbol{x}_i,\boldsymbol{y}_t)|
\end{align}
}%
where we use the definition 1 in (13), basic arithmetic in (14) and the property of the exponent function in (15). Using the triangle inequality in metric space $(\mathbb{R}^{d_s},d)$ , we know that

{\small
\begin{align}
    d(\boldsymbol{x}_i,\boldsymbol{z}_t) - d(\boldsymbol{y}_t,\boldsymbol{z}_t) \leq d(\boldsymbol{x}_i,\boldsymbol{y}_t) \leq d(\boldsymbol{x}_i,\boldsymbol{z}_t) + d(\boldsymbol{y}_t,\boldsymbol{z}_t)
\end{align}
}%

hence 

{\small
\begin{align}
    k|f(\boldsymbol{x}_i)-f(\boldsymbol{x}_j)|
    &< \sum_{1\leq t \leq k} |d(\boldsymbol{x}_j,\boldsymbol{z}_t)-d(\boldsymbol{x}_i,\boldsymbol{y}_t)|\\
    & \leq \sum_{1\leq t \leq k} \left( |d(\boldsymbol{x}_j,\boldsymbol{z}_t)-d(\boldsymbol{x}_i,\boldsymbol{z}_t)| + d(\boldsymbol{y}_t,\boldsymbol{z}_t) \right)\\
    & \leq \sum_{1\leq t \leq k} d(\boldsymbol{x}_i,\boldsymbol{x}_j) + \sum_{1\leq t \leq k} d(\boldsymbol{y}_t,\boldsymbol{z}_t)\\
    & \leq k \cdot d_{ij} + k \cdot d_{ij}\\
    & = 2 k d_{ij}
\end{align}
}%

Note that we apply (16) in (18), the triangle inequality and Definition 1 in (19). Finally, we have $|f(\boldsymbol{x}_i)-f(\boldsymbol{x}_j)| / d_{ij} < 2$.


\quad

In summary, we can find a constant $c=\max\{1/a,2\}$, such that $|f(\boldsymbol{x}_i)-f(\boldsymbol{x}_j)|/d_{ij} \leq c$.

\clearpage
\subsection{Metrics}
\label{sup:metrics}

\begin{table}[h]
    \centering
    \caption{Important symbols used in the metric definition.}
    \resizebox{\columnwidth}{!}{
    \begin{tabular}{cc}
    \toprule
         Symbol & Description \\ \midrule
         $Prec_i$ & Precision of class $i$ \\
         $Rec_i$ & Recall of class $i$\\
         $TP_i$ & The number of true positive samples of class $i$\\
         $FP_i$ & The number of false positive samples of class $i$\\
         $FN_i$ & The number of false negtive samples of class $i$\\
         $F_i$ & The F1-score of class $i$\\
         $Fscore$ & The weighted F1-score overall classes\\
         $C_i$ & The $i$-th class\\
         $n$ & The number of samples. $\sum_{i}{|C_i|}=n$\\
    \bottomrule
    \end{tabular}
    }
    \label{tab:symbols_appendix}
\end{table}

\paragraph{F1-score.} The F1-score is a measure of a model’s accuracy on a dataset. It is the harmonic mean of precision and recall for each class. In the multi-class case, the overall metric is the average F1 score of each class weighted by support (the number of true instances for each label). The higher F1-score, the better the result. The mathematical definition can be found in Eq.~\ref{eq: F1_score}.
\begin{equation}
    \begin{cases}
        Fscore = \frac{1}{n} \sum_{i} |C_i| \cdot F_i;\\
        F_i = 2 \frac{Prec_i \cdot Rec_i}{ Prec_i+Rec_i };\\
        Prec_i = \frac{TP_i}{TP_i+FP_i};\\
        Rec_i = \frac{TP_i}{TP_i+FN_i}.\\
    \end{cases}
    \label{eq: F1_score}
\end{equation}

\paragraph{ARI.} The Rand Index computes a similarity measure between two clusterings by considering all pairs of samples and counting pairs that are assigned in the same or different clusters in the predicted and true clusterings:
\begin{equation}
    RI = \frac{TP+TN}{TP+TN+FP+FN},
    \label{eq: RI}
\end{equation}
where $TP$ is the number of true positives, $TN$ is the number of true negatives, $FP$ is the number of false positives, and $FN$ is the number of false negatives. The raw RI score is then “adjusted for chance” into the ARI score using the following scheme:
\begin{equation}
    ARI = \frac{RI-\mathbb{E}[RI]}{\max(RI)-\mathbb{E}(RI)}.
    \label{eq: ARI}
\end{equation}
The higher the ARI, the better the clusterings. By introducing a contingency Table.~\ref{tab:contingency}, the original Adjusted Rand Index value is:
\begin{equation*}
    ARI = \frac{ \sum_{ij}\binom{n_{ij}}{2} - [\sum_i \binom{a_{i}}{2} \sum_j \binom{b_j}{2}]/ \binom{n}{2} }{[\sum_i \binom{a_i}{2}+\sum_j \binom{b_j}{2}]/2 - [\sum_i \binom{a_i}{2} \sum_j \binom{b_j}{2}]/\binom{n}{2} }.
\end{equation*}

\begin{table}[h]
    \centering
    \caption{The contingency table. Given a set of $n$ elements, and two partitions (e.g. clusterings) of these elements, namely $X=\{X_1,X_2,\ldots,X_r\}$ and $Y=\{ Y_1,Y_2,\ldots,Y_s \}$, the overlap between $X$ and $Y$ can be summarized in a contingency table $[n_{ij}]$ where each entry $n_{ij}$ denotes the number of objects in common between $X_i$ and $Y_j$: $n_{ij}=|X_i \cap Y_j|$.}
    \begin{tabular}{c|cccc|c}
               & $Y_1$ & $Y_2$ & $\dotso$ & $Y_s$ & sums \\ \hline               
        $X_1$  & $n_{11}$ & $n_{12}$ & $\dotso$ & $n_{1s}$ & $a_1$ \\
        $X_2$  & $n_{21}$ & $n_{22}$ & $\dotso$ & $n_{2s}$ & $a_2$ \\
        $\vdots$  & $\vdots$ & $\vdots$ & $\ddots$ & $\vdots$ & $\vdots$ \\
        $X_r$  & $n_{r1}$ & $n_{r2}$ & $\dotso$ & $n_{rs}$ & $a_r$ \\ \hline
        sums   & $b_1$ & $b_2$ & $\dotso$ & $b_s$ &  \\
    \end{tabular}
    \label{tab:contingency}
\end{table}

\paragraph{NMI.} Normalized Mutual Information (NMI) is a normalization of the Mutual Information (MI) score to scale the results between 0 (no mutual information) and 1 (perfect correlation). Using notations in Table.~\ref{tab:contingency}, NMI can be computed by \cite{geng2018recome}:
\begin{equation}
    NMI = \frac{ \sum_{i=1}^r\sum_{j=1}^s{\frac{n_{ij}}{n}\log{\frac{n n_{ij}}{a_i b_j}}} }{ \sum_{i=1}^r\frac{a_i}{n}\log{\frac{a_i}{n}} \sum_{j=1}^s \frac{b_j}{n}\log{\frac{b_j}{n}} }.
\end{equation}

\clearpage
\subsection{Pseudocode}

\begin{algorithm}
    \caption{Local cluster detection algorithm \\
            \textbf{Complexity}: $\mathcal{O}(k d_s n\log{n})$}
    \label{algorithm: local_cluster_detecting}
    \begin{algorithmic}[1]
        \Require sample set $\mathcal{X}$, neighborhood size $k$; 
        \Ensure local clusters $V$, boundary pair $\mathcal{B}$, intensity $f(\boldsymbol{x})$;
        \State {\color{blue} Init}: $\hat{r}_i \gets \boldsymbol{x}_i$ for each $i$, where $\hat{r}_i$ is the root index of $\boldsymbol{x}_i$
        
        \State
        \State {\color{blue} Step 1}: Compute neighborhood and intensity for each point  (line 4-5).
        \For{ $i \gets 0, n-1$}
            \State Calculate $\mathcal{N}_{i}$ and $f(\boldsymbol{x_i})$, Eq.\ref{eq:intensity} \Comment{{\color{gray}$\mathcal{O}(k d_s n\log{n})$}}
        \EndFor
        
        \State
       \State {\color{blue} Step 2}: Find root for each point (line 10-14). Record boundary pair between various local clusters (line 15-17).
       
        \State $idx \gets \arg$ sort$(\{-f(\boldsymbol{x}_i)\}_{i=0}^{n-1})$ \Comment{{\color{gray}$\mathcal{O}(n\log{n})$}}
        
        \While{$idx \neq \phi$}  \Comment{ {\color{gray}$\mathcal{O}(n k^2)$} }
            \State $i \gets idx[0],\lambda \gets f(\boldsymbol{x}_i)$
            \State $\mathcal{J} \gets \{j|j \in \mathcal{N}_i, f(\boldsymbol{x}_i) \geq \lambda \}$ \Comment{{\color{gray} $\mathcal{O}(k)$}}

            \State $j \gets \arg\max_{j\in \mathcal{J}} \frac{f(\boldsymbol{x}_j)-f(\boldsymbol{x}_i)}{||\boldsymbol{x}_i-\boldsymbol{x}_j||}$, Eq. \ref{eq:Prx} \Comment{{\color{gray}$\mathcal{O}(k)$}}
            
            \If {$j \neq \phi$} \Comment{{\color{gray}parent exists}}
                \State $\hat{r}_i \gets \hat{r}_j$  \Comment{{\color{gray}update root}}
                \For{$s \in \mathcal{N}_i$} \Comment{{\color{gray}record $\mathcal{B}$, $\mathcal{O}(k^2)$}}
                    \If{$\hat{r}_s \neq \hat{r}_i$ and $i \in \mathcal{N}_s$}
                        \State $\mathcal{B}$.append($(i,s,\hat{r}_i,\hat{r}_s)$) 
                    \EndIf
                \EndFor
            \EndIf
            
            \State idx.remove($i$) 
        \EndWhile
        
        \State
        \State {\color{blue} Step 3}: Collect points for local clusters (line 21-23)
        \State $V=\{\}$ \Comment{{\color{gray} init an empty set}}
        \For{ $i \gets 0, n-1$}  \Comment{{\color{gray}$\mathcal{O}(n)$}}
            \State $V[r_i]$.append($i$) 
        \EndFor
    \end{algorithmic}
    
    Note that $\mathcal{N}_i$ is the set of neighborhoods' index of $\boldsymbol{x}_i$.
\end{algorithm}

\begin{algorithm}
    \caption{Topo-graph construction \\
            \textbf{Complexity}: $\mathcal{O}(kn)$}
    \label{algorithm: topo-graph construction and pruning}
    \begin{algorithmic}[1]
    \Require boundary pairs $\mathcal{B}$, intensity $f(\boldsymbol{x})$, threshold $\alpha$
    \Ensure set of edge $E$
        \State {\color{blue} Init}: $ E_{i,j}=0$ for all $i,j$
        
        \State
        \State {\color{blue} Step 1}: Construct topo-graph (line 4-6)
        \For{$(i,j,\hat{r}_i,\hat{r}_j) \in \mathcal{B}$} \Comment{{\color{gray}$\mathcal{O}(|\mathcal{B}|)<\mathcal{O}(kn)$}}
            \State $s \gets \frac{(f(\boldsymbol{x}_i)+f(\boldsymbol{x}_j))^2}{4|v_{r_i}| \cdot |v_{r_j}|}$, Eq.\ref{eq: distance_local_clusters2} and Eq.\ref{eq: point_distance}.
            \State $E_{\hat{r}_i,\hat{r}_j}+=s,E_{\hat{r}_j,\hat{r}_i} \gets E_{\hat{r}_i,\hat{r}_j}$
        \EndFor
        
    \end{algorithmic}
\end{algorithm}

\begin{algorithm}
    \caption{Topo-graph pruning (connectivity)\\
            \textbf{Complexity}: $\min \{ \mathcal{O}(|E||V|\log{|V|}),\mathcal{O}(|E|\log{|E|}) \}$}
    \label{algorithm: topo-graph construction and pruning}
    \begin{algorithmic}[1]
    \Require local clusters $V$, edge $E$, class number $\boldsymbol{n}=[n_0,n_1,\ldots,n_c]$
    \Ensure the label mapping of each local clusters $C[-1]$
        \State {\color{blue} Init}: Initialization (line 2-8)
        \State Create a list $\mathcal{E}=[(i,j,E_{i,j})  \text{for  all}  E_{i,j}>0] $, such that $\mathcal{E}_{k,2} \geq \mathcal{E}_{k+1,2}$ \Comment{{\color{gray}$\mathcal{O}(|E|\log{|E|})$}}
        \State $C =[\{ i:i \ \mathrm{for} \ i \ \mathrm{in} \ V.keys()\}]$ \Comment{{\color{gray} node index $\mapsto$ class index}}
        \State $M=[\{ i:[i] \ \mathrm{for} \ i \ \mathrm{in} \ V.keys()\}]$
        \Comment{{\color{gray} class index $\mapsto$ member node indexes}}
        \State $N=[\{ i:\mathrm{len}(V[i]) \ \mathrm{for} \ i \ \mathrm{in} \ V.keys() \}]$ \Comment{{\color{gray} node index $\mapsto$ number of nodes of the same class}}
        \State score = [9999] \Comment{{\color{gray} intermediate dissimilarity}}
        \State c = len($\boldsymbol{n}$)
        
        \State
        \State {\color{blue} Step 1}: Merge nodes by the descending order of edge strength, generate intermediate class ratios seq[:,c], get the dissimilarity sc, and determine whether accept this merge operation (line 11-28)
        \For{ $i,j,v$ in $\mathcal{E}$} 
            \State $C_{prev}=C[-1]$
            \State $M_{prev}=M[-1]$
            \State $N_{prev}=N[-1]$
            \If{$C_{prev}[i] \neq C_{prev}[j]$} \Comment{{\color{gray} merge different classes}}
                
                \State $S_i = M_{prev}[i]; S_j = M_{prev}[j]$
                \State $S = M_{prev}[i] + M_{prev}[j]$
                \State Create $C_{now}$ by Eq.~\ref{eq: update_C} for all $k$. \Comment{{\color{gray} $\mathcal{O}(|V|)$}}
                \State Create $M_{now}$ by Eq.~\ref{eq: update_M} for all $k$. \Comment{{\color{gray} $\mathcal{O}(|V|)$}}
                \State Create $N_{now}$ by Eq.~\ref{eq: update_N} for all $k$. \Comment{{\color{gray} $\mathcal{O}(|V|)$}}
                
                \State seq = sort( [$N_{now}[c]$ for $c$ in set($C_{now}$.values())] ), descending order. \Comment{{\color{gray} $\mathcal{O}(|V| \log{|V|})$}}
                \If{$len(seq) < c$}
                    \State break
                \EndIf
                
                \State sc = $S(C_{now},\boldsymbol{\rho})$
                
                \If{ sc $\leq$ score[-1] } \Comment{{\color{gray} if accept merging}}
                    \State score.append(sc)
                    \State C.append($C_{now}$)
                    \State M.append($M_{now}$)
                    \State N.append($N_{now}$)
                \EndIf
                

            \EndIf
        \EndFor
    

    \end{algorithmic}
\end{algorithm}

\paragraph{Auxiliary functions.}
During the process of topo-graph pruning (Algorithm 3), we use the following auxiliary functions for efficiency concerns:

(1) Mapping local cluster index to final cluster index
\begin{equation}
    \hat{C}[k] = 
    \begin{cases}
       C[i] & k \in S_j\\
       C[k] & else
    \end{cases}
    \label{eq: update_C}
\end{equation}

(2) Mapping final cluster index to local cluster index
\begin{equation}
    \hat{M}[k] = 
    \begin{cases}
       S & k \in S\\
       M[k] & else
    \end{cases}
    \label{eq: update_M}
\end{equation}

(3) Recording the number of samples for each final class:
\begin{equation}
    \hat{N}[k] = 
    \begin{cases}
       N[i]+N[j] & k \in S\\
       N[k] & else
    \end{cases}
    \label{eq: update_N}
\end{equation}

\clearpage
\subsection{Sensitive analysis}
\paragraph{Objective and Setting.} As mentioned before, GIT has one hyperparameter $k$ which needs to be tuned. To study how sensitive the results are to this hyperparameter, we change $k$ from 30 to 100, and compare GIT with $k$-mean++ on MNIST and FMNIST.

\begin{figure}[h]
    \centering
    \includegraphics[width=2.5in]{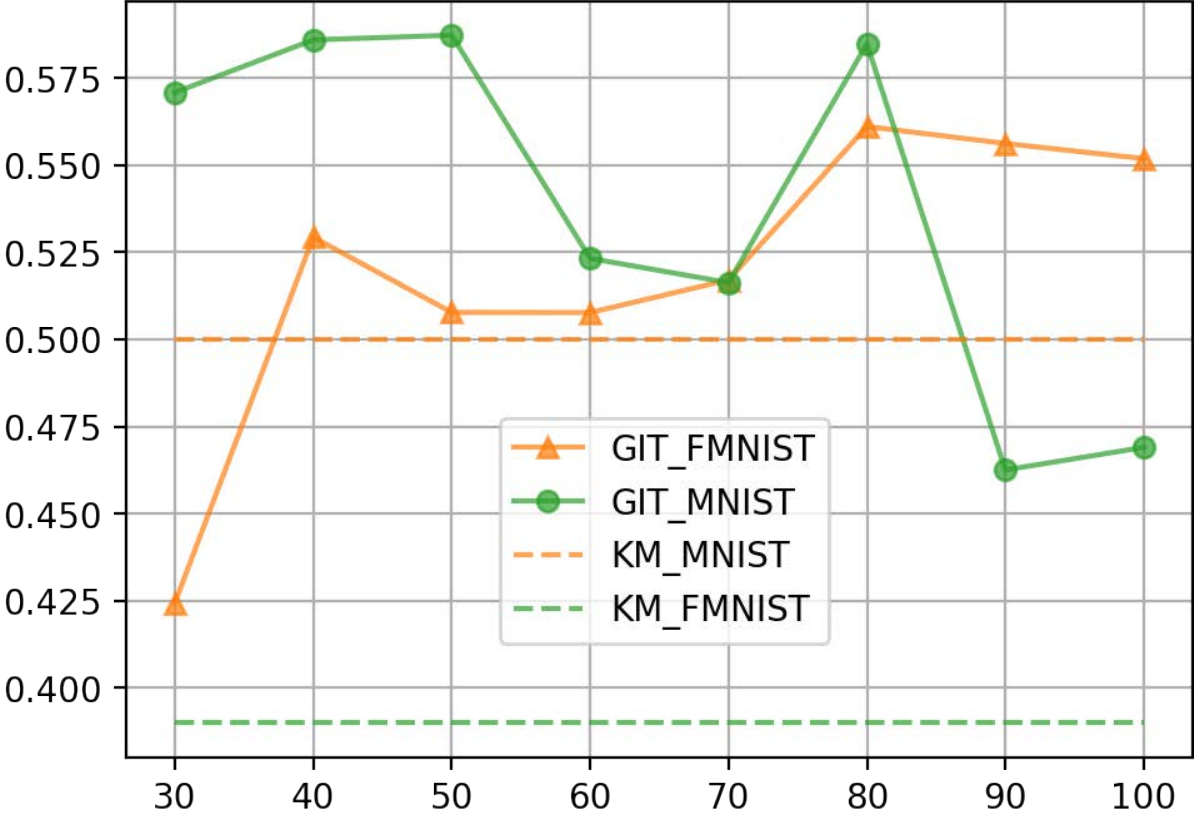}
    \caption{ F1-score of GIT on the raw dataset (dim=784) under different $k$. The x-axis and y-axis are $k$ and F1-score, respectively. Dashed lines represent baseline results of $k$-means++, and solid lines represent results of GIT. Results of different data sets are colored differently. }
    \label{fig: raw_data }
\end{figure}

\begin{figure}[h]
    \centering
    \includegraphics[width=3.4in]{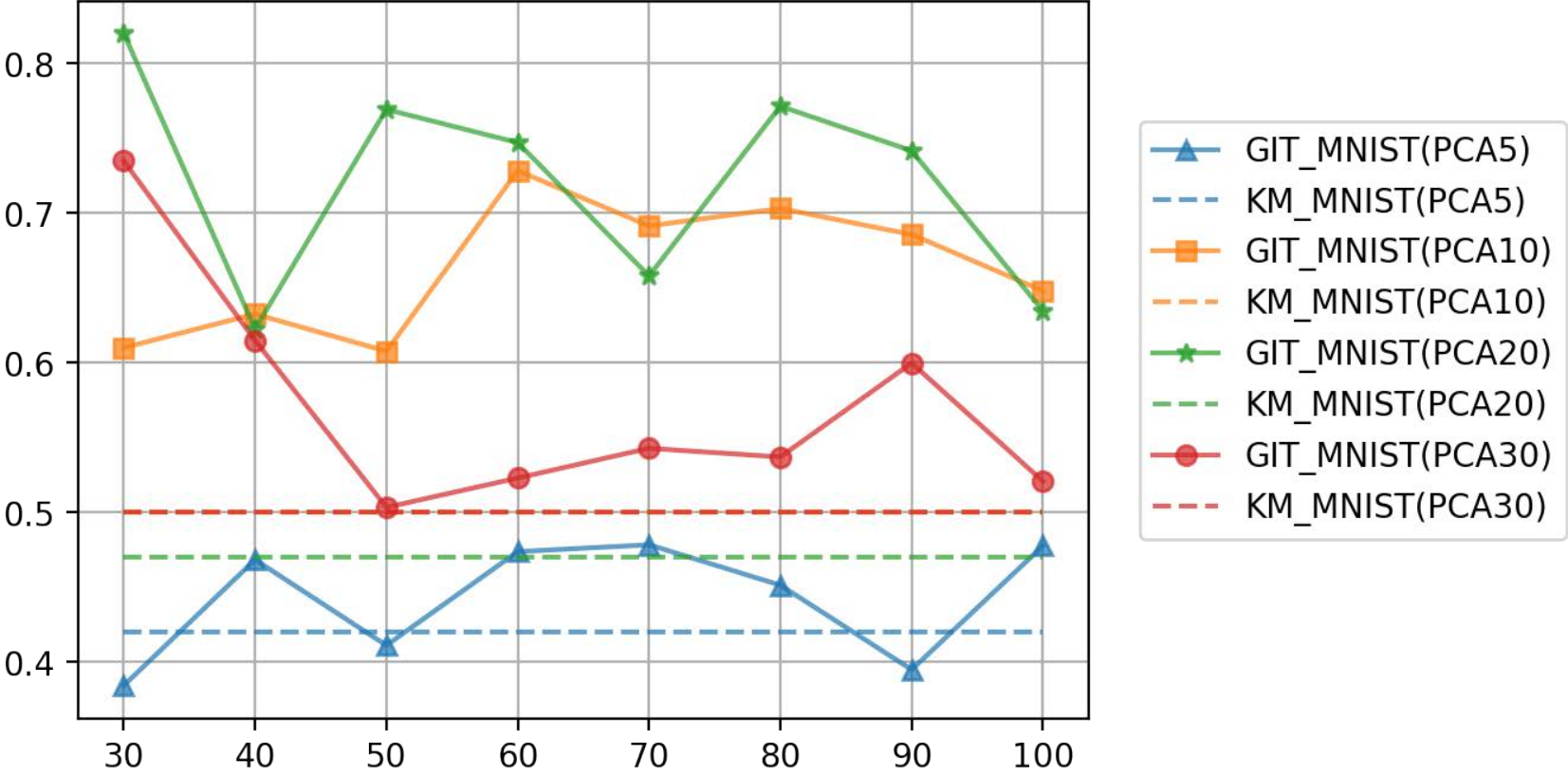}
    \caption{ PCA+MNIST, changing $k$. The x-axis and y-axis are $k$ and F1-score. \textit{GIT\_MNIST(PCA5)} indicates that we use PCA to project the original MNIST data into a $5$-dimensional space, and then reported GIT's accuracy on it. }
    \label{fig:PCA_MNIST}
\end{figure}

\begin{figure}[H]
    \centering
    \includegraphics[width=3.4in]{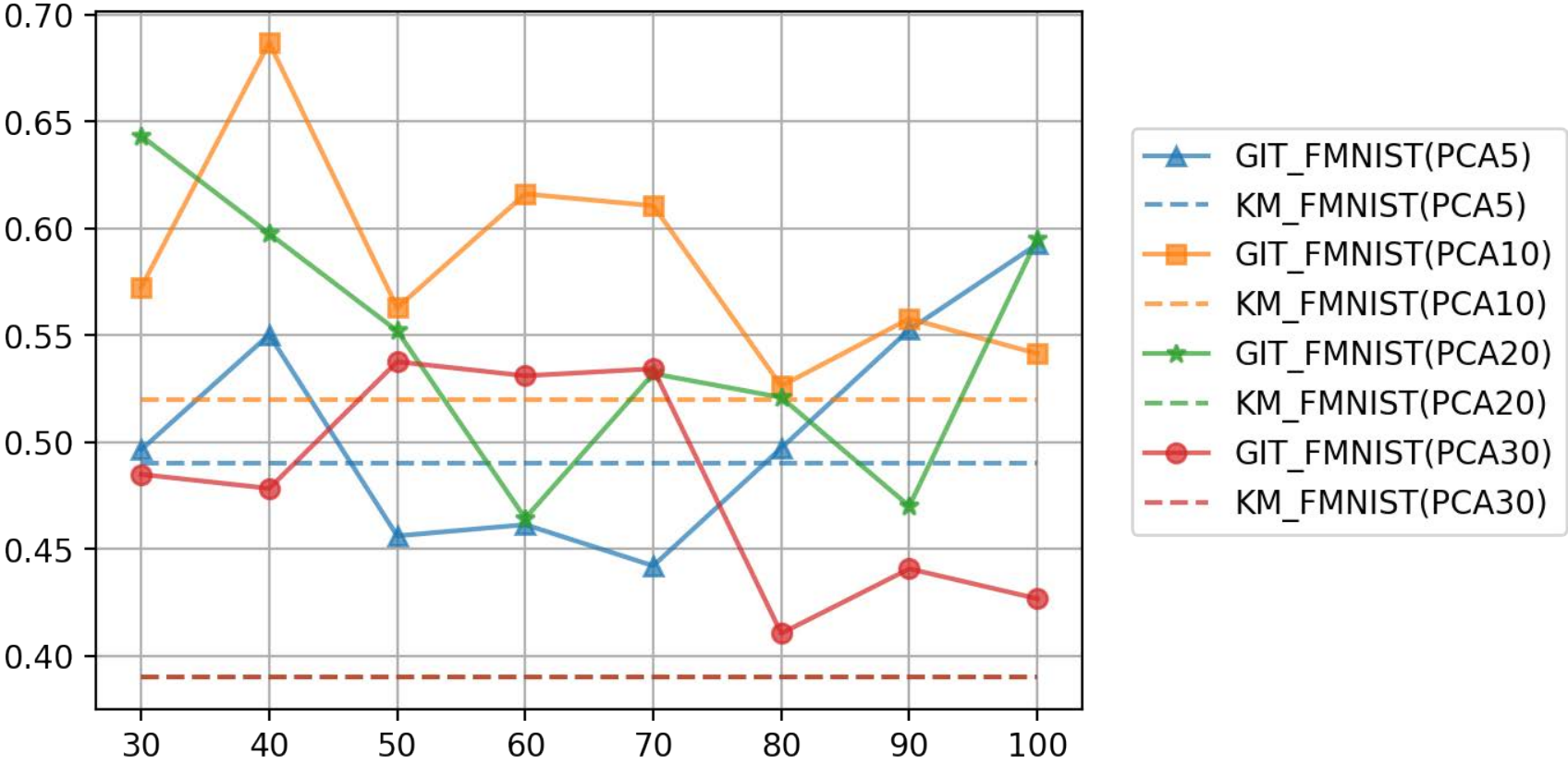}
    \caption{ PCA+FMNIST, changing $k$. }
    \label{fig:PCA_FMNIST}
\end{figure}

\begin{figure}[h]
    \centering
    \includegraphics[width=3.4in]{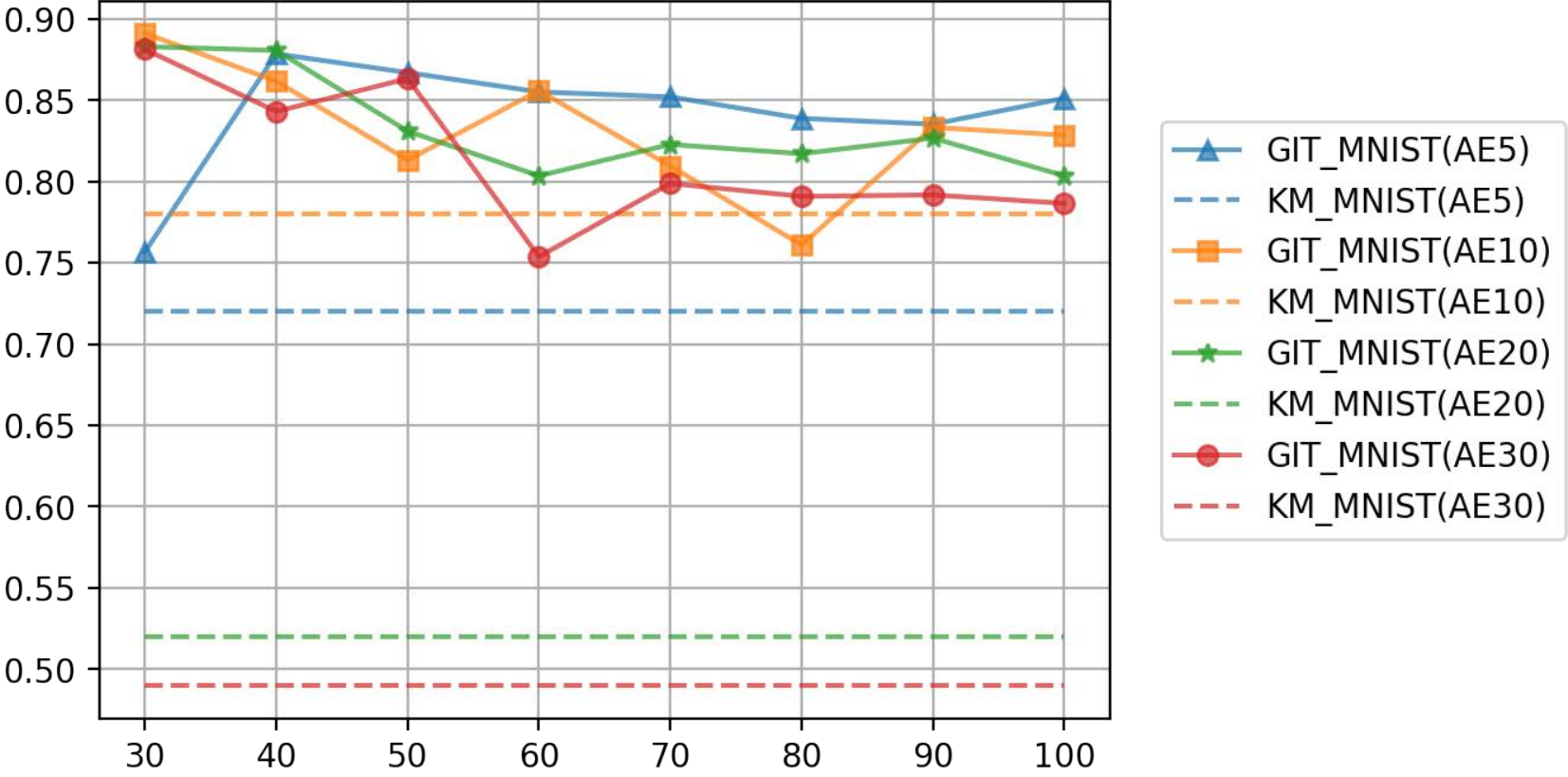}
    \caption{ AE MNIST, changing $k$. }
    \label{fig:AE_MNIST}
\end{figure}

\begin{figure}[h]
    \centering
    \includegraphics[width=3.4in]{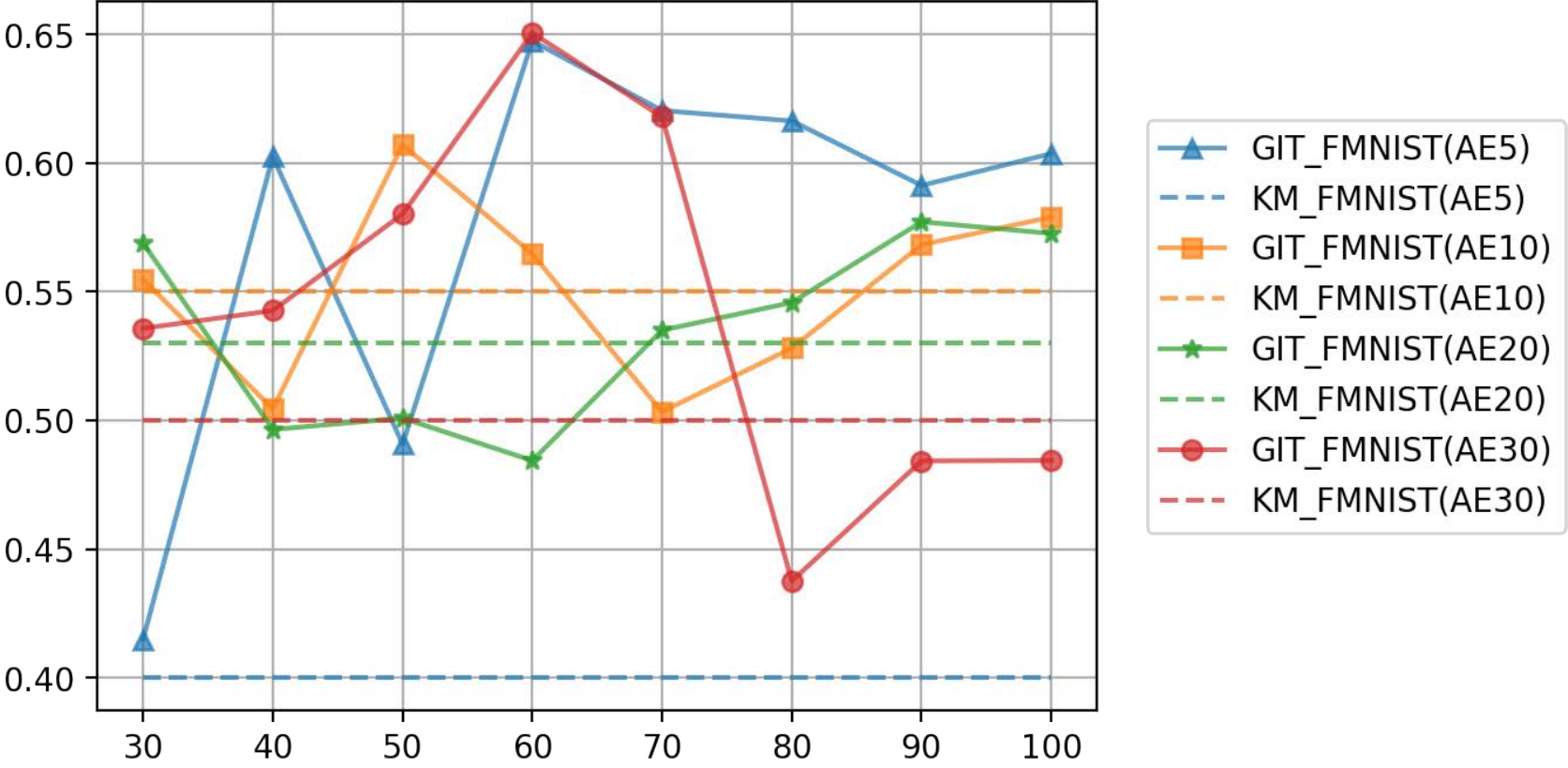}
    \caption{ AE FMNIST, changing $k$. }
    \label{fig:AE_FMNIST}
\end{figure}

\paragraph{Results and Analysis.} Generally, we cannot ensure the results of GIT are robust to $k$. However, we can claim that GIT can significantly outperform the baseline method in the vast majority of cases. Thanks to this property, it is easy to select a proper parameter for better performance.

\clearpage
\subsection{Potential questions}
\paragraph{Q1: Do the authors use labels to tune hyperparameters?}

\paragraph{R1:} Yes, but it is fair and reasonable for all algorithms. Concretely, we perform clustering from artificially specified hyperparameters and evaluate the F1-score using labels after clustering. For each algorithm, we repeat this process to find better results. We do this because we think it is not reasonable to report results derived from randomly selected hyperparameters for different algorithms with different hyperparameters. Thus, we use some labels to guide hyperparameter searching and fairly report the best results that we can find for each algorithm.

\paragraph{Q2: In Table.~\ref{tab:shapes}, why different baselines are selected for different data sets? }

\paragraph{R2:} There is some misunderstanding here. We evaluated all baselines mentioned in this paper, but only visualize part of them in Table.~\ref{tab:shapes} due to page limitations. More specifically, we choose results with top-3 F1-score for visualization.

\paragraph{Q3: How do the authors treat 'uncovered' points in metrics computation?}

\paragraph{R3:} All the metrics are calculated based on 'covered' points without considering 'uncovered' points. If the cover rate is less than 80\%, the corresponding results will be ignored and marked in gray.

\paragraph{Q4: How do the authors visualize results using UMAP in Figure.~\ref{fig:visualization}? } 

\paragraph{R4:} We use UMAP to project original data (dimension=5) into 2-dimensional space for visualization convenience. Then, we color each point with real label (ground truth) or a predicted label (generated from a clustering algorithm). Due to the information loss caused by dimension reduction, points in different classes may overlap.

\paragraph{Q5: Why not report averagy results under different random seeds?}

\paragraph{R5:} There are three cases:
\begin{itemize}[itemindent=1em]
    \item When adding noises to study the robustness, we report the average results due to the varying noises.
    \item As to accuracy, GIT is deterministic which means the accuracy does not fluctunate with the change of random seeds under the fixed hyperparameters. Thus, there is no need to report average results of different seeds.
    \item As to speed, the reported running time is close to the average based on our experimental experience. Since the running time of different algorithms varies greatly and random seed will not cause the change of the magnitude order, our results are sufficient to distinguish them.
\end{itemize}

\paragraph{Q6: Why different baselines are chosen in different experiments? Do authors prioritize favorable baselines?}

\paragraph{R6:} We don't prioritize favorable baselines because that would be cheating. We choose classical $k$-means++ and Spectral Clustering as they are widely used. By comparing GIT with them, readers can extend their understanding to a wider range of situations. We also choose recent HDBSCAN \cite{mcinnes2017accelerated}, Quickshift++ \cite{jiang2018quickshift++}, SpectACl \cite{hess2019spectacl} and DPA \cite{d2021automatic} as SOTA methods. In each experiment, we have evaluated all baseline algorithms along with GIT. However, we cannot present all the results due to the page limitations, although we would like to. As a compromise, we present the results of the most representative and effective algorithms in each experiment.

If you have carefully read this paper, you would discover that we:
\begin{itemize}[itemindent=1em]
    \item compare \{DPA, FSF, HDB, QSP, SA\} with GIT in Table.~\ref{table:GIT_density_small} (accuracy on small-scale datasets) $\Rightarrow$ select the most competitve SOTA methods \{HDB,QSP,SA\} for further comparison.
    \item compare \{KM, SC\} $\cup$ \{HDB,QSP,SA\} with GIT in Table.~\ref{table:GIT_density_large} (accuracy on large-scale datasets) $\Rightarrow$ the most accurate classical algorithm is \{KM\} and the top-2 accurate SOTA algorithms are \{QSP,SA\}.
    \item compare \{QSP,SA\} with GIT in terms of robustness in Table.~\ref{tab:noises_scales} and Fig.~\ref{fig:robustness_change}.
    \item compare \{KM,QSP,SA\} with GIT along with dimension reduction in Fig.~\ref{fig:visualization}.
    \item compare all algorithms in Table.~\ref{tab:shapes}.
\end{itemize}

The above is the logic of our experiments. We only show the most representative results and ignore others due to page limitations. Perhaps we may miss some interesting open source baselines, but we are willing to provide further comparisons. Thanks for your reading.

\end{document}